\documentclass[sigconf]{acmart}

\usepackage{url}
\usepackage{epsfig,epstopdf,algorithmic,algorithm,amsfonts,url,textcomp,amsthm,multirow,moresize}
\usepackage{graphicx,enumitem}
\usepackage{amsmath}\allowdisplaybreaks
\usepackage{amsmath}
\usepackage{booktabs}

\usepackage{graphicx}
\usepackage{amsmath}
\usepackage{booktabs}
\usepackage{algorithm}
\usepackage{algorithmic}
\usepackage{amsmath, bm}
\usepackage{float}
\usepackage{multicol}
\usepackage{multirow}
\usepackage{float}
\usepackage{subfigure}
\usepackage{dsfont}

\def\BibTeX{{\rm B\kern-.05em{\sc i\kern-.025em b}\kern-.08emT\kern-.1667em\lower.7ex\hbox{E}\kern-.125emX}}

\copyrightyear{2020}
\acmYear{2020}
\setcopyright{acmlicensed}
\acmConference[Woodstock '18]{Woodstock '18: ACM Symposium on Neural Gaze Detection}{June 03--05, 2018}{Woodstock, NY}
\acmBooktitle{Woodstock '18: ACM Symposium on Neural Gaze Detection, June 03--05, 2018, Woodstock, NY}
\acmPrice{15.00}
\acmDOI{10.1145/1122445.1122456}
\acmISBN{978-1-4503-9999-9/18/06}

\begin{document}

%
\title{
	Reinforcement Learning with Supervision from Noisy Demonstrations
}

%
%
%
%
\author{Kun-Peng Ning}
\affiliation{%
	\institution{Nanjing University of Aeronautics and Astronautics}
	\city{Nanjing}
	\country{China}}
\email{ningkp@nuaa.edu.cn}

\author{Sheng-Jun Huang}
\affiliation{%
  \institution{Nanjing University of Aeronautics and Astronautics}
  \city{Nanjing}
  \country{China}}
\email{huangsj@nuaa.edu.cn}
%
%
%

%

%
\begin{abstract}
    Reinforcement learning has achieved great success in various applications. To learn an effective policy for the agent, it usually requires a huge amount of data by interacting with the environment, which could be computational costly and time consuming. To overcome this challenge, the framework called Reinforcement Learning with Expert Demonstrations (RLED) was proposed to exploit the supervision from expert demonstrations. Although the RLED methods can reduce the number of learning iterations, they usually assume the demonstrations are perfect, and thus may be seriously misled by the noisy demonstrations in real applications. In this paper, we propose a novel framework to adaptively learn the policy by jointly interacting with the environment and exploiting the expert demonstrations. Specifically, for each step of the demonstration trajectory, we form an instance, and define a joint loss function to simultaneously maximize the expected reward and minimize the difference between agent behaviors and demonstrations. Most importantly, by calculating the expected gain of the value function, we assign each instance with a weight to estimate its potential utility, and thus can emphasize the more helpful demonstrations while filter out noisy ones. Experimental results in various environments with multiple popular reinforcement learning algorithms show that the proposed approach can learn robustly with noisy demonstrations, and achieve higher performance in fewer iterations.





\end{abstract}

%
%


%

%

%
\maketitle
	\section{Introduction}
	\label{sec:intro}
	In recent years, reinforcement learning (RL) has achieved significant progress as a method of building intelligent agents for decision making. The target of reinforcement learning is to find a policy to decide the optimal action given a state, such that the future reward received from the environment is maximized~\cite{szepesvari2010algorithms,kaelbling1996reinforcement}.
	However, RL algorithms usually require a huge number of interactions with the environment to learn an effective policy. For example, in Atari Games, 44 million frames are used to train the agent with model-free deep Q-learning~\cite{hessel2018rainbow}; in Go game, strategic policies that combined with search cost 40 days to learn the model~\cite{silver2016mastering, silver2017mastering}; in Autonomous Driving, the agent attempts more than 310 thousand times to learn to park~\cite{schulman2017proximal}. Obviously, such large scale interactions can be applied only in simulated environment, while remain impractical in most real world applications.
	
	In fact, in many tasks, we may have some demonstration trajectories from experts as the supervised information, which can significantly reduce the number of interactions required for learning the policy. Reinforcement learning with expert demonstrations (RLED) is a framework to exploit such supervised information. The key idea of RLED is that RL algorithms can save many experiences by incorporating prior knowledge of various forms into the learning process. These methods usually work in a two-step manner, i.e., firstly pre-train through supervised learning from demonstrations and then learn policy by exploring the environment. For example, some methods combine an imitation hinge loss with the Q-learning loss to minimize the optimal Bellman residual~\cite{hester2018deep,piot2014boosted}, the Approximate Policy Iteration method use expert demonstrations to define linear constraints of the optimization problem~\cite{kim2013learning}, and in ~\cite{chemali2015direct}, a classification-based policy iteration algorithm is proposed to imitate the expert policy.
	These methods typically assume the demonstrations are perfect, and their ultimate goal is to derive suitable behaviors from the demonstrations. Unfortunately, in real cases, demonstrations usually contain serious noises or even misleading information. As a result, the policy learned from the noisy demonstrations is less likely to be consistent with the ground-truth target, and thus may fail when applied in real tasks.
	
	In this paper, instead of assuming that the demonstrations are perfect, we consider a more practical setting, where the demonstrations could be noisy or imperfect. We propose to learn the policy with the supervision from noisy demonstration (LfND for short). The basic idea is that, different demonstrations may have different effects to the policy learning, and even the same demonstration may contributes differently at different time. If we can accurately estimate the potential utility of each demonstration at each iteration, the supervised information can be exploited more adequately.

To implement this idea, we firstly form an instance for each step of the demonstration trajectory, which is a triplet of state, action and reward $(s,a,r)$. Then, we define a joint loss function on the instance to let the agent simultaneously interact with the environment and exploit the expert demonstrations. On one hand, policy gradient is performed to maximize the expected reward; on the other hand, a cross entropy loss is defined to minimize the distance between the actions of the agent and the demonstration. Finally, we assign each instance with a weight based on the difference between the state-action value $Q_E(s,a)$ calculated from expert demonstration and the state value $V_{\pi}(s)$ estimated by the state value function. This difference can be regarded as the expected gain of the value function for a specific instance, and thus measures its potential contribution to the policy learning. In other words, the noise or misleading demonstrations will be filtered out with small weights, while the useful demonstrations will be emphasized with large weights. By minimizing the weighted loss function over all instances, the demonstrations are fully exploited as a supervision to the environment exploration.

	
	We implemented our approach with multiple popular reinforcement learning algorithms, and perform experiments in different environments. The results show that the policy learned by our method can lead to better performance in most cases. Especially the proposed method can robustly exploit the noisy demonstrations to significantly reduce the number of training iterations.

	The rest of this paper is organized as follows. We review related work in Section \ref{section.related_work} and introduce the proposed method in Section \ref{section.proposed}. Section \ref{section.experiments} reports the experiments, followed by the conclusion in Section \ref{section.conclusion}.


	\section{Related Work}
	\label{section.related_work}

	In reinforcement learning, expert demonstrations have been shown to contribute effectively in challenging environments~\cite{subramanian2016exploration}. Learning from demonstrations (LfD) \cite{schaal1997learning}, which tries to clone the behavior from the demonstrations, is a typical framework to exploit the experts' knowledge. Imitation learning is one of such approaches that directly trains an action predictor from demonstration data~\cite{schaal1997learning,atkeson1997robot}. Recently, some studies propose to explore an adversarial paradigm for the behavior cloning method~\cite{ho2016generative, wu2019imitation}. Another popular paradigm is inverse reinforcement learning (IRL)~\cite{ng2000algorithms,abbeel2004apprenticeship,ziebart2008maximum}, which expected to find a proper reward function that can explain the demonstrations as the optimal behavior. Some methods tried to shape reward using demonstrations or expert advice, which expected that the learned reward is more effective for agents to explore the environment~\cite{brys2015reinforcement,suay2016learning}.



Recently, more approaches try to directly learn with demonstration trajectories~\cite{piot2014boosted}. Most of these methods focus on optimizing TD loss based on Q-learning models~\cite{hester2018deep,piot2014boosted,kim2013learning,chemali2015direct}, and thus are less suitable for problems with continuous action space. For example, DQfD~\cite{hester2018deep} and OBR~\cite{piot2014boosted} are two similar methods, which perform imitation and exploration by combining TD loss and the classification loss. The process of AlphaGo~\cite{silver2016mastering} works in a different way to combine exploration and demonstration based on policy gradient algorithm. Specifically, AlphaGo firstly pre-trains a policy network from a dataset of 30 million expert state-action pairs, and then uses this network as a initialization to train policy net by applying policy gradient methods. Another approach to combine RL and demonstration is POfD~\cite{kang2018policy}, which also focuses on the policy-based method. Specifically, the POfD approach defines a Jensen-Shannon (JS) divergence between the learned policy and expert's policy, and optimizes it through adversarial training on demonstrations.

A common assumption of the above methods is that the expert demonstrations are noise-free, which can hardly hold in real environments. Recently, there is one work called NAC~\cite{gao2018reinforcement} noticed the potential risk of imperfect demonstrations. The NAC approach defined an extra regular term to normalize the Q-function, which can reduce the Q-values of actions unseen in the demonstration data. However, this method focuses on improving the robustness of the model by regularization, and does not explicitly estimate the different contributions of demonstrations with different qualities. 
	
	\section{The LfND Approach}
	\label{section.proposed}
	In this section, we first formalize the framework for reinforcement learning with noisy demonstrations, and then introduce the proposed LfND approach in detail.
	
	\subsection{Problem Setting}
	We consider an agent interacting with the environment over a sequence of steps, which can be formalized as a Markov Decision Process $\mathcal{M}=(\mathcal{S},\mathcal{A},\gamma,\mathcal{P},\mathcal{R})$. Here $\mathcal{S}$ is a set of states, $\mathcal{A}$ is a set of actions, $\gamma \in [0,1)$ is the disount factor, $\mathcal{P}$ is the state transition probabilities, and $\mathcal{R}:(\mathcal{S}\times\mathcal{A}) \rightarrow \mathcal{R}$ is the reward function.
	At each step $t$, with the state $s_t\in\mathcal{S}$, the agent takes an action $a_t\in\mathcal{A}$ according to the policy $\pi$, and receives a reward $r_t\in\mathcal{R}$ from the environment. The target is to maximize the discounted sum of rewards over all steps.

	\begin{figure*}[htbp]
		\centering
		\subfigure[Existing RLED methods]{
			\label{fig.pre}
			\includegraphics[width=0.46\textwidth]{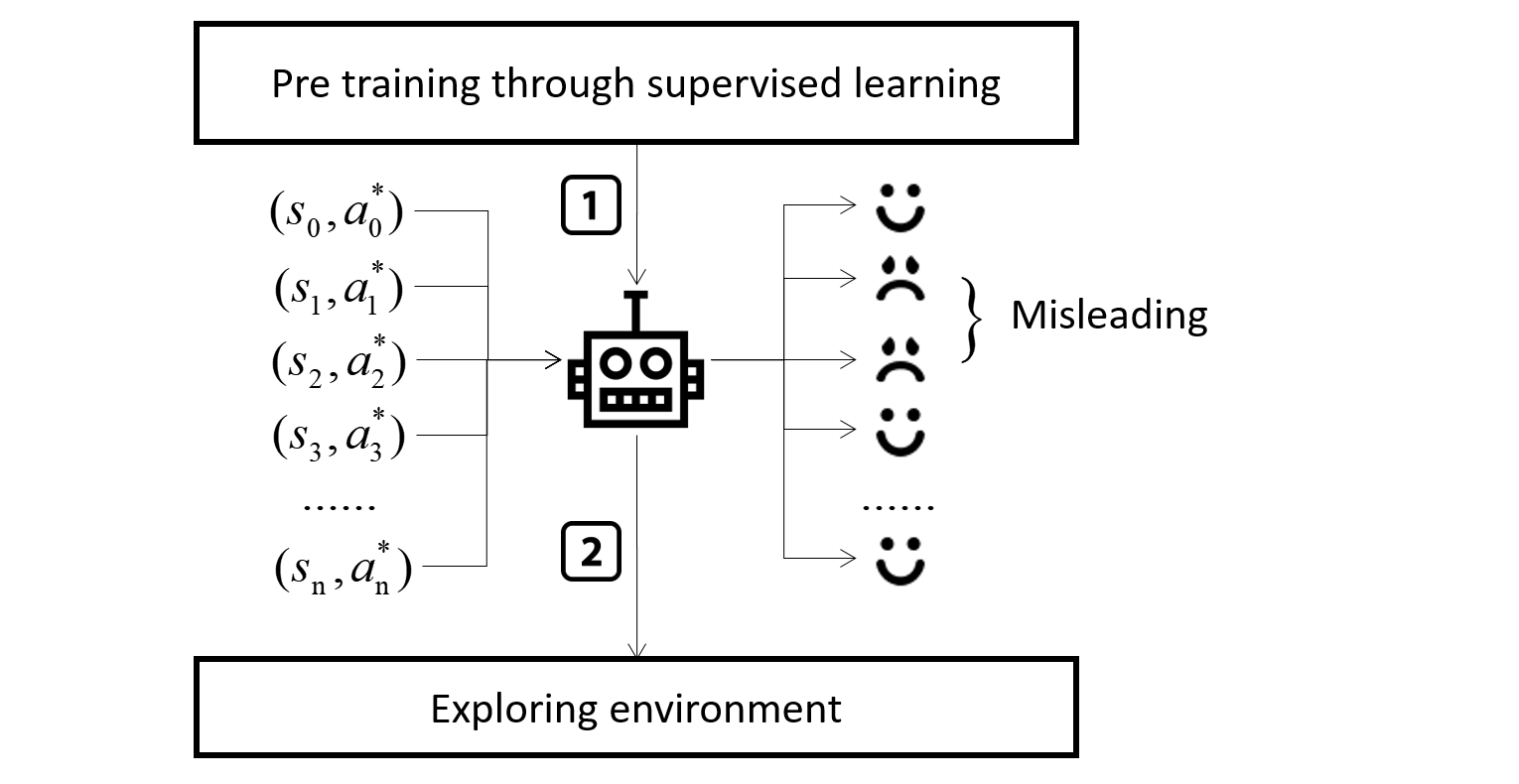}}\qquad\quad
		\subfigure[the proposed LfND framework]{
			\label{fig.now}
			\includegraphics[width=0.46\textwidth]{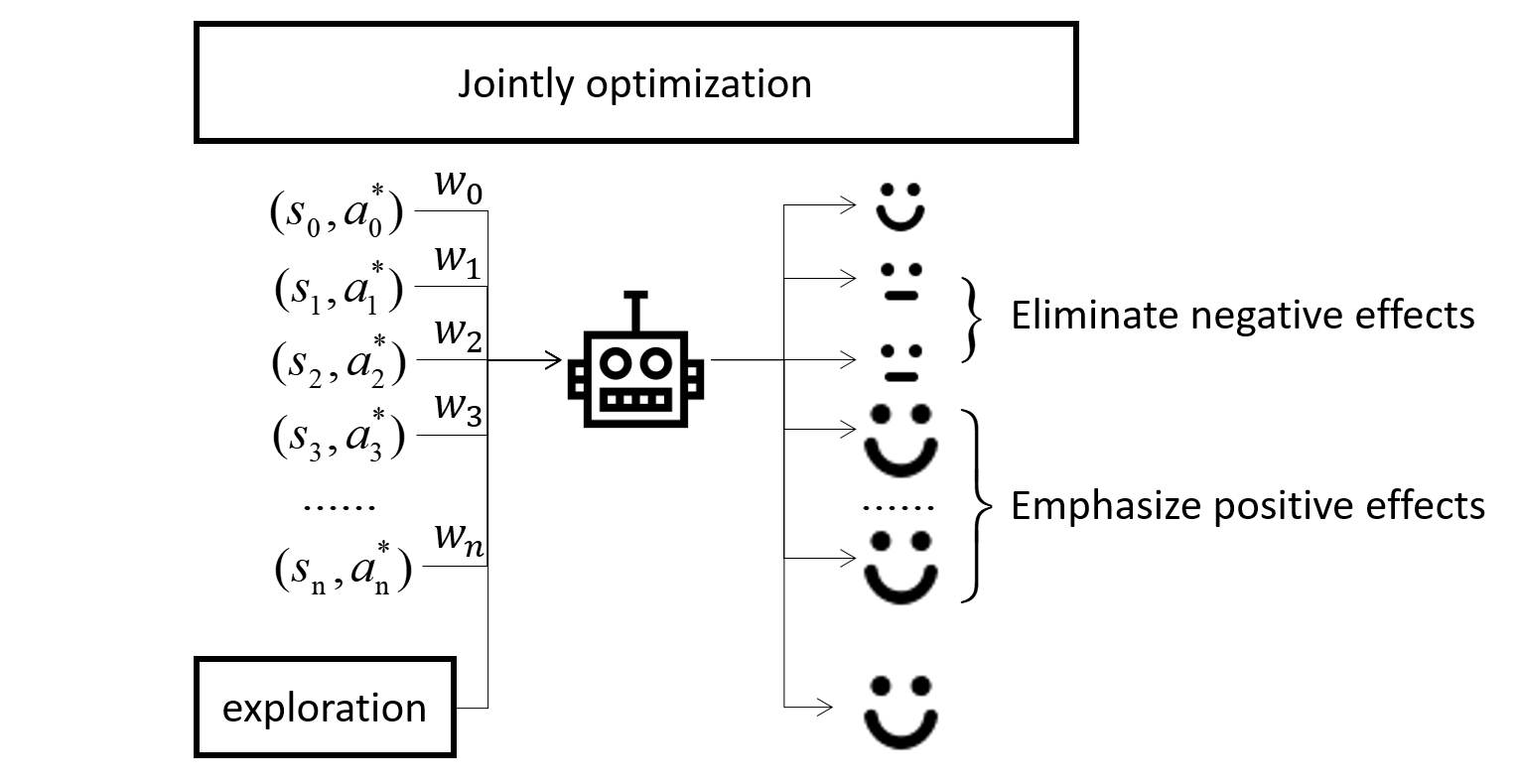}}
		\caption{Comparison on the framework of the proposed lfND approach and previous RLED methods.}
		\label{fig.comparison}
	\end{figure*}

	RL algorithms typically learn an effective control policy after many millions of steps, which is unacceptable in most real tasks. To overcome this problem, we can provide the agent with expert demonstrations to more effectively learn the policy $\pi(\cdot)$. Formally, a trajectory is a sequence of observations, actions and rewards, $\sigma=((s_1,{a^*}_1,r_1),(s_2,{a^*}_2,r_2),...,(s_T,{a^*}_T,r_T))$, where ${a^*}_i$ is the $i$-th action from experts, and $s_T$ is the terminal state. For convenience, we denote $(s_i,{a^*}_i)$ as an instance corresponding to the $i$-th step of the trajectory $\sigma$.

As discussed in the Introduction, the trajectory $\sigma$ collected from expert demonstration is usually imperfect, because it may contain serious noise or even some misleading information. For example, to train an agent to play Go game with human demonstrations, we should be aware that a trajectory win the game does not mean that it is optimal. The optimal demonstrations are hard to obtain caused by data collection noise or produced by the immaturity of the expert. It is more common that some parts of the demonstrations are optimal while the others are not. A naive approach to handle noisy demonstrations is to simply filter out the trajectories with low reward. However, a low reward does not necessarily imply that all the demonstration steps in the trajectory are useless. Instead, even for a noisy trajectory, some steps of it may still provide important information for the policy optimization. Moreover, even for the same step of a trajectory, it may contribute differently at different stages of the learning process, as the model changes.

Based on the above observations, we propose to adaptively estimate the potential utility of each instance at each iteration during the policy learning. Formally, we introduce a weight variable $w_i$ for each demonstration instance $(s_i,{a^*}_i)$ to estimate the potential value. Obviously, more useful instances should receive higher values of $w_i$, while noisy instance should receive lower values of $w_i$. After that, we try to learn the policy $\pi(\cdot)$ by allowing the agent to interact with the environment and exploit the supervised information from weighted demonstration instances.


	\subsection{Algorithm Detail}	
	First of all, we compare the proposed framework with existing reinforcement learning with expert demonstrations (RLED) methods in Figure \ref{fig.comparison}. The framework of previous approaches~\cite{gao2018reinforcement,piot2014boosted,kim2013learning,chemali2015direct,silver2016mastering} are demonstrated in Figure \ref{fig.pre}. They typically work in a two-steps way. In the first step, a supervised learning algorithm is employed to pre train the policy network based on the demonstrations. Then in the second step, the policy is optimized by allowing the agent to explore the environment. Obviously, the demonstration learning and environment exploring are performed separately, making it less possible to adaptively exploit the supervised information from the demonstration trajectories. More importantly, all the instances of the trajectories are uniformly used, without considering their potential utility. When the demonstration trajectories are imperfect, some noisy instances will mislead the policy optimization, and subsequently hurt the learning performance.
	
In contrast, the proposed LfND framework is demonstrated in Figure \ref{fig.now}. First of all, it simultaneously learns from the demonstrations and the environment in a joint framework, which allows the agent to adaptively utilize the information from two sources. Unlike previous methods that assume the demonstrations are perfect, the proposed LfND framework introduces a weight variables $w_i$ for each instance $(s_i, a^*_i)$ to estimate its potential utility. Noticing that because the utility of a specific instance varies as the model changes, the weights are adaptively updated in different learning iterations. By optimizing the weighted loss over all instances, it is expected to emphasize the positive effects of good demonstrations while eliminate the negative effects of noisy demonstrations.

	In the following part of this section, we will firstly introduce the joint training of the policy $\pi(\cdot)$ based on the noisy demonstrations and environment exploring, and then discuss on how to calculate the weights to accurately estimate the potential utility of demonstration instances.

To allow joint training of the agent by simultaneously learning from the demonstrations and exploring the environment, we define a joint objective function as in Eq. \ref{Eq:loss2}.
	\begin{equation}\label{Eq:loss2}
	\ell = \ell_d+ \lambda \ell_e,
	\end{equation}
where $\ell_d$ and $\ell_e$ denote the loss for demonstration learning and environment exploring, respectively, and $\lambda$ is a trade-off parameter.

For demonstration learning, as previously discussed, given a set of noisy trajectories, if we can estimate the potential utility of each instance accurately at each time, it is more effective to utilize these supervised information for policy joint optimization. In other words, our target is to utilize the right demonstration instances at the right time to learn an effective policy. Specifically, in the set of demonstration trajectories $\Sigma = \{\sigma^1,\sigma^2,...,\sigma^m\}$, each trajectory $\sigma^i$ consists of $n^i$ (state, action, reward) instances as follows: $$\sigma^i=((s^i_1,{a^*}^i_1,r^i_1),(s^i_2,{a^*}^i_2,r^i_2),...,(s^i_{n^i},{a^*}^i_{n^i},r^i_{n^i})).$$
	Given a state $s^i_j$, based on the currently learned parameters $\theta$ of the policy network, the agent will return an action:
	$$a_j^i=\pi_{\theta}(s_j^i).$$
	Note that we assume both $a^i_j$ and ${a^*}^i_j$ is a distribution over all actions, where each element describes the probability of taken the corresponding action. Then, to learn the policy $\pi(\cdot)$ from demonstration, we try to minimize the distance between $a^i_j$ and ${a^*}^i_j$, forcing the agent to behave similarly to the expert. Formally, we define a weighted cross-entropy loss function over all trajectories as Eq. \ref{Eq:loss1}.
	\begin{equation}\label{Eq:loss1}
	\ell_d=\sum_{i=1}^m\sum_{j=1}^{n_i}-w_j^i\cdot {{a^*}_j^i}\log{a_j^i}.
	\end{equation}
	Obviously, by minimizing the loss function, the policy $\pi$ will be optimized to produce consistent actions with the experts, while the negative effect of noisy demonstrations will be eliminated by the weight $w_j^i$. Note that we only introduce this straightforward implementation to validate our idea of learning from demonstrations. Other more advanced strategies may be employed to further improve the performance.

For environment exploring, we focus on policy gradient methods, which are popular and can handle tasks with continuous actions compared to Q-learning methods. As an example, we can employ the trust region policy optimization (TRPO) method~\cite{schulman2015trust} to define the loss function $\ell_e$ as follows:
	\begin{equation}\label{Eq:trpo}
	\begin{split}
	\ell_e(\theta)=-\hat{\mathds{E}}[\frac{\pi(\bm{a|s})}{\pi^{old}(\bm{a|s})}\hat{A}] \,,\\
	s.t.\quad \hat{\mathds{E}}[KL[\pi^{old}(\bm{\cdot|s}),\pi(\bm{\cdot|s})]] \le \beta,
	\end{split}
	\end{equation}
	where $\pi^{old}$ is the policy parameters before the update, $\hat{A}$ is an estimator of the advantage function, and the $\beta$ is the parameters of max trust region between the new policy and the old policy.
	
	Alternatively, one can also use proximal policy optimization (PPO) algorithm~\cite{schulman2017proximal} to define $\ell_e$ as follows:
	\begin{equation}\label{Eq:ppo}
	\ell_e(\theta)=-\hat{\mathds{E}}[min(r_t(\theta)\hat{A},clip(r_t(\theta),1-\epsilon,1+\epsilon)\hat{A})].
	\end{equation}
	where $r_t(\theta)$ denote the probability ratio $r_t(\theta) = \frac{\pi(\bm{a|s})}{\pi^{old}(\bm{a|s})}$. The second term of Eq~\ref{Eq:ppo}, $clip(r_t(\theta),1-\epsilon,1+\epsilon)\hat{A}$, modifies the surrogate objective by clipping the probability ratio, which prevents a big gap between the new and old policies~\cite{schulman2017proximal}.

Finally, by substituting Eqs. \ref{Eq:loss1} and \ref{Eq:trpo} or \ref{Eq:ppo} into Eq. \ref{Eq:loss2}, we get the final objective function of the proposed LfND approach. It can be observed that both the two terms of objective function are consistently trying to learn the policy distribution, and can be effectively solved with policy gradient methods.

Next, we discuss the estimation of the potential utility of each instance, i.e., calculating the weights $w_j^i$ for each instance $(s^i_j, {a^*}^i_j)$. Intuitively, given a state $s_j^i$, if the action given by the expert will lead to a higher expected reward than the currently learned policy, it is likely that the expert policy is superior to the agent policy, and thus such demonstration could be utilized to improve the policy network. We thus define the weight as follows:
    \begin{equation}\label{Eq:wei}
	\begin{split}
	w_j^i = \mathds{1}{\{Q_{\sigma^i}(s_j^i,{a^*}_j^i) - V_\pi(s_j^i) \geq 0\}}
	\end{split}
	\end{equation}
	where
	\begin{equation*}
	\begin{split}
	&Q_{\sigma^i}(s_j^i,{a^*}_j^i) = \sum_{k=j}^{n_i}\gamma^{k-j} r_j^i\,,
	\end{split}
	\end{equation*}
	Here, $\mathds{1}{\{\cdot\}}$ is the indicator function that the value is 1 when the condition is met, otherwise it is 0. $Q_{\sigma^i}(s_j^i,{a^*}_j^i)$ is the state-action value function calculated from expert, which estimates the long-term reward of performing action ${a^*}_j^i$ in state $s_j^i$. Note that this term is an accurate value because the whole trajectory $\sigma^i$ is given. $V_\pi(s_j^i)$ is the state value function from policy $\pi$, which estimates the value of the state $s_j^i$. It can be observed Eq. \ref{Eq:wei} has a similar form with the advantage function $A_\pi(s,a)=Q_\pi(s,a)-V_\pi(s)$~\cite{baird1993advantage,wang2015dueling}. The difference is that advantage function aims at describing how good it is to select action $a$ in state $s$ under the current policy $\pi$, and its $Q_\pi$ function is approximated by policy $\pi$.

\textbf{Remarks:} As we discussed before, the utility of a specific demonstration may vary over different learning stages as the model updates. For example, at the begining, the parameter of policy network $\theta$ is usually initialized to a small number that makes the $V_\pi(s_j^i)$ is also a small number. Thus nearly all instances will be used to optimize the policy $\pi$ because the $Q_{\sigma^i}(s_j^i,{a^*}_j^i) - V_\pi(s_j^i)$ will be greater than zero for most instances. As the learning process goes on, the $V_\pi(\cdot)$ will increase, and the weights of some less useful instances will be zero so that these instances will not participate in the policy optimization. Note that, it can not work if we use advantage function $Q_\pi(s,a)-V_\pi(s)$ instead of $Q_{\sigma^i}(s_j^i,{a^*}_j^i) - V_\pi(s_j^i)$,	because $Q_\pi$ and $V_\pi$ are both approximated by policy $\pi$, which can not estimate weights of instances.

Another case is that even the demonstration $(s_j^i,{a^*}_j^i)$ is noise, it is still possible to contribute to the policy optimizatioin. Intuitively, this instance may have a negative effect on optimizing a policy with high performance, but it still has a positive effect on the randomly initialized policy or a policy with poor performance. This also demonstrate that the proposed approach can dynamically adjust the weights through the current policy $\pi$, and adaptively utilize the right instances at right time to jointly optimize the policy.
	
	
	
	
	\begin{algorithm}[tb]
		\caption{The LfND algorithm}
		\label{alg:LfND}
		\begin{algorithmic}[1]
			\STATE \textbf{Input:}
			\STATE \quad Environment $\mathcal{E}$;
			\STATE \quad Observation Space $\mathcal{O}$;
			\STATE \quad Action Space $\mathcal{A}$;
			\STATE \quad Set of trajectory $\Sigma=\{\sigma^1,\sigma^2,...,\sigma^m\}$;
			\STATE \textbf{Process:}
			\STATE \quad Initialize $\theta$ randomly;
			\STATE \quad Initialize memory $\mathcal{H}=\emptyset$;
			\STATE \textbf{Repeat:}
			\STATE \quad Obtain state $s_0$ from the environment $\mathcal{E}$.
			\STATE \quad Choose $a_0 \sim \pi_\theta(s_0)$.
			\STATE \quad $t=1$;
			
			\STATE \quad \textbf{Repeat:}
			\STATE \quad \quad Execute action $a_{t-1}$ and observe reward $r_t$ and
			\STATE \quad \quad next state $s_t$;
			\STATE \quad \quad Store tuple $(s_{t-1}, a_{t-1}, r_t, s_t)$ in $\mathcal{H}$;
			\STATE \quad \quad Choose $a_t \sim \pi(s_t)$;
			\STATE \quad \quad $t = t + 1$;
			\STATE \quad \textbf{until} the end of this round
			
			\STATE \quad Caculate $w_j^i$ for each instance by Eq.~\ref{Eq:weight_deta} or Eq.~\ref{Eq:weight_log};
			\STATE \quad Update $\theta$ by minimizing $\ell$ in Eq.~\ref{Eq:loss2} from $\Sigma$ and $\mathcal{H}$;
			\STATE \quad $\mathcal{H}=\emptyset$;
			
			\STATE \textbf{until} convergence or desirable performance
			\STATE Output the learned policy $\pi$.
			
		\end{algorithmic}
	\end{algorithm}

	\begin{figure*}[!ht]
		\centering
		\subfigure[HalfCheetah]{
			\label{fig.HalfCheetah}
			\includegraphics[width=0.15\textwidth]{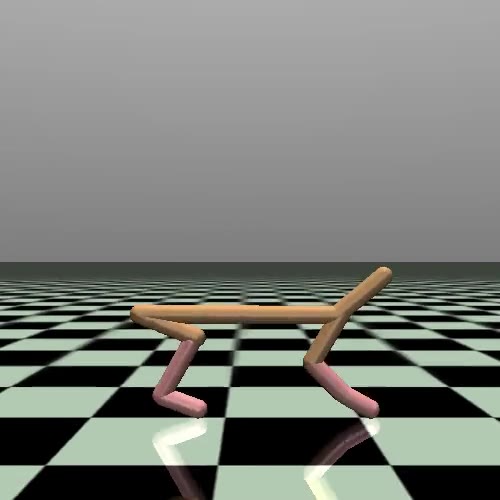}}
		\subfigure[Hopper]{
			\label{fig.Hopper}
			\includegraphics[width=0.15\textwidth]{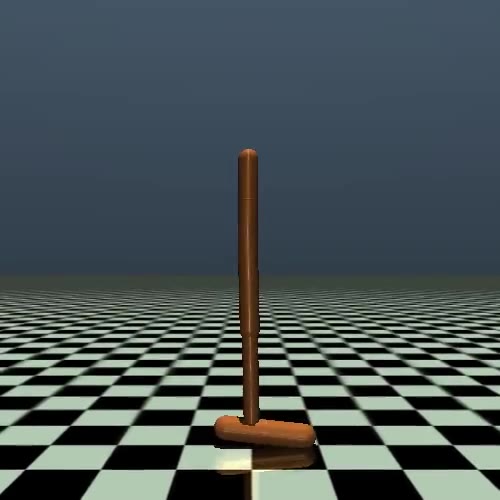}}
		\subfigure[Humanoid]{
			\label{fig.Humanoid}
			\includegraphics[width=0.15\textwidth]{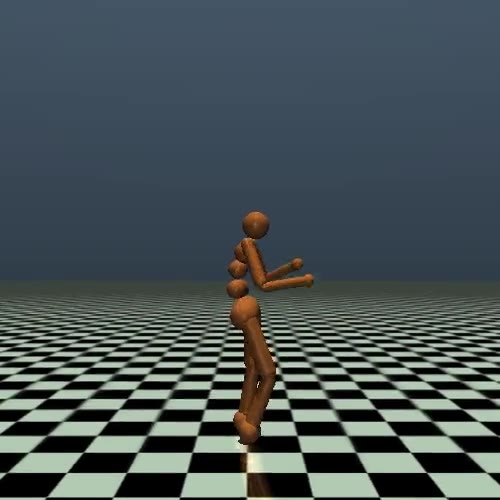}}
		\subfigure[HumanoidStandup]{
			\label{fig.HumanoidStandup}
			\includegraphics[width=0.15\textwidth]{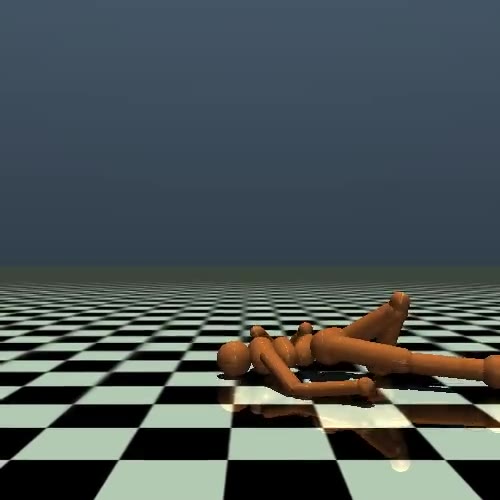}}
		\subfigure[InvertedPendulum]{
			\label{fig.InvertedPendulum}
			\includegraphics[width=0.15\textwidth]{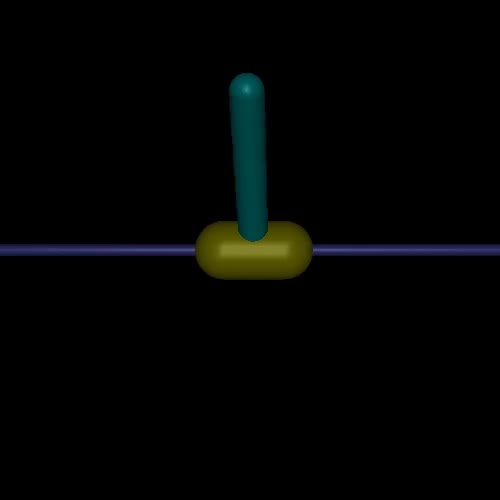}}
		\subfigure[Walker2d]{
			\label{fig.Walker2d}
			\includegraphics[width=0.15\textwidth]{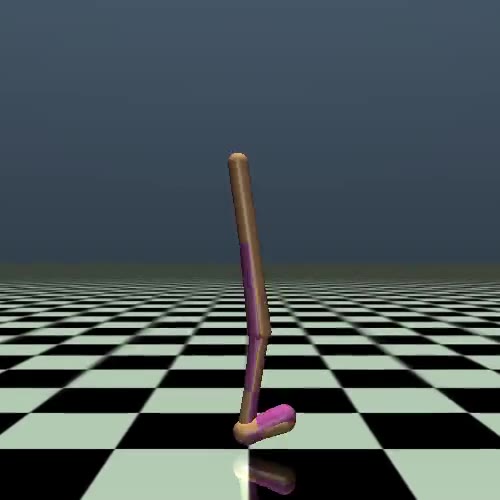}}
		\caption{Screen shots of the Mujoco environments used in the experiments.}
		\label{fig.clf}
	\end{figure*}

	Although the weight defined in Eq. \ref{Eq:wei} can eliminate the negative effects of noisy demonstrations, it is unable to distinguish demonstrations with different degrees of positive effects. Therefore, we provide some other optional solutions to the weight estimation. Specifically, we define a linear form of weight as shown in Eq~\ref{Eq:weight_deta}.
	\begin{equation}\label{Eq:weight_deta}
	w_j^i = max{\{\frac{Q_{\sigma^i}(s_j^i,{a^*}_j^i) - V_\pi(s_j^i)}{\delta} ,0\}}.
	\end{equation}
	where $\delta$ is a hyperparameter, say, $\delta=10$.

Similarly, we also define a logarithmic form of the weight as shown in Eq~\ref{Eq:weight_log}.
	\begin{equation}\label{Eq:weight_log}
	w_j^i = log(max{\{Q_{\sigma^i}(s_j^i,{a^*}_j^i) - V_\pi(s_j^i) ,1\}}).
	\end{equation}
	Obviously, Eq~\ref{Eq:weight_deta} and Eq~\ref{Eq:weight_log} can better describe the importance of different demonstration instances. The comparison among these different implementations will be studied in the experiment section.
	
	The process of the approach is summarized in Algorithm \ref{alg:LfND}. Firstly, environment $\mathcal{E}$, observation space $\mathcal{O}$, action space $\mathcal{A}$ and set of trajectory $\Sigma$ are given. Then the parameters $\theta$ represents the policy network is randomly initialized. We also introduce the memory $\mathcal{H}$ to store tuples of agent interactions, which is initialized as an empty set.
	At each iteration, some trajectories generated by interaction between agent and environment are stored in memory $\mathcal{H}$, and then $w_j^i$ for each demonstration instance will be caculated by Eq.~\ref{Eq:weight_deta} or Eq.~\ref{Eq:weight_log}, finally $\theta$ will be optimized until convergence or desirable performance. When updating $\theta$ by minimizing the loss function $\ell$ in Eq.~\ref{Eq:loss2}, $\ell_e(\theta)$ can be solved with existing policy gradient algorithms.

	
%
%
	
	
	


	\section{Experiments}
	\label{section.experiments}
	\subsection{Settings}
	To validate the effectiveness of the proposed approach, we perform experiments in six commonly used MuJoCo~\cite{todorov2012mujoco} environments, which are implemented in OpenAI Gym~\cite{brockman2016openai}.
	\textbf{HalfCheetah:} a simulated cheetah robot system with 17-dimensional state and 3-dimensional action, whose goal is to make the robot run as far as possible.
	\textbf{Hopper:} a one-legged robot, which is expected to hop forward as fast as possible in this environment. There is a 11-dimensional state space and a 3-dimensional action space.
	\textbf{Humanoid:} there is a bipedal robot in the Humanoid system with 376-dimensional state and 17-dimensional action. Its goal is to make it walk forward as fast as possible without falling over.
	\textbf{HumanoidStandup:} the HumanoidStandup system with 376-dimensional state and 17-dimensional action is trying to make the robot standing up as long as possible.
	\textbf{InvertedPendulum:} the goal of this environment is to prevent an agent which moves along a frictionless track from falling over. In this environment, a 4-dimensional state and a 1-dimensional continuous action is provided.
	\textbf{Walker2d:} the goal of Walker2d system is to make the robot run as fast as possible. The state is in 17-dimensional and the action is in 6-dimensional.
Screen shots of the six environments are shown in Figure~\ref{fig.clf}. For each environment, 10 demonstration trajectories are provided by a well trained agent, among which 5 trajectories contain noisy demonstrations. 
Specifically, all trajectories are generated by the trained agent, and each trajectory is from the initial state to the terminal state, where the noisy demonstrations are produced by the immature agent.

	We respectively employ two state-of-the-art reinforcement learning algorithms, i.e., TRPO~\cite{schulman2015trust} and PPO~\cite{schulman2017proximal}, as the base model to implement our approach as well as other compared methods. The following methods are compared in the experiments:
	\begin{itemize}
		\item \textbf{Trust Region Policy Optimization (TRPO):} A RL method performs strategy search in the trust region, and tries to ensure that the strategy of the next iteration will be better than the current strategy~\cite{schulman2015trust}.
		
		\item \textbf{Proximal Policy Optimization (PPO):} A RL method that tries to optimize the lower bound of the clipped and unclipped objectives~\cite{schulman2017proximal}.
		
		\item \textbf{Imitation Learning only (IL):} Supervised imitation from expert demonstrations without any environment interaction.
		
		\item \textbf{Learning policy by AlphaGo (LbA):} This method combines exploration and demonstration based on policy gradient method as in AlphaGo. Specifically, it firstly pre trains the policy network with expert demonstrations by supervised learning, and then trains the network by exploring the environment~\cite{silver2016mastering}. 
		
		\item \textbf{Policy Optimization from Demonstration (POfD):} A latest RLED method that optimizes the JensenShannon (JS) divergence between the current policy and expert's policy by adversarial training on demonstrations~\cite{kang2018policy}.
				
		\item \textbf{LfND:} The approach proposed in this paper. %

		\item \textbf{LfND without weight (LfND-noW):} A degenerated version of the proposed method, which uses all the instances of the trajectories uniformly in the learning process.
	\end{itemize}
	The parameters of the two base models are set as recommended in the corresponding literatures. Specifically, for TRPO, the penalty coefficient $\beta$ is set to 0.01. For PPO, the parameter $\epsilon$ is set to 0.2. We use the same parameter setting for both our approach and the compared methods. The trade-off parameter $\lambda$ of LfND is set to 1 as default. We use linear form of weight in experiments and set $\delta=10$ as default.

	\begin{figure*}[!ht]
		\centering
		\subfigure[HalfCheetah]{
			\label{fig.TRPO-HalfCheetah}
			\includegraphics[width=0.32\textwidth]{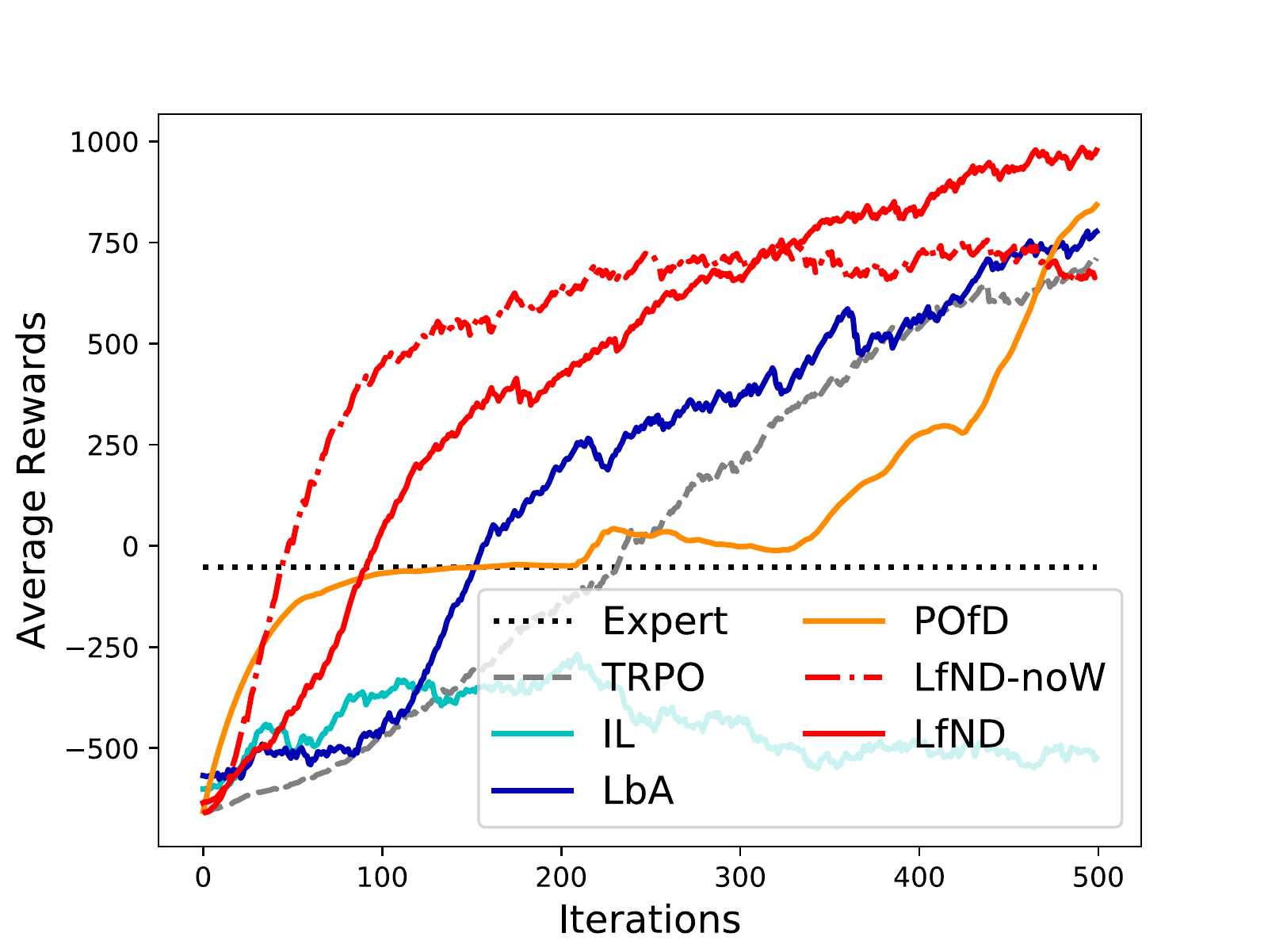}  }
		\subfigure[Hopper]{
			\label{fig.TRPO-Hopper}
			\includegraphics[width=0.32\textwidth]{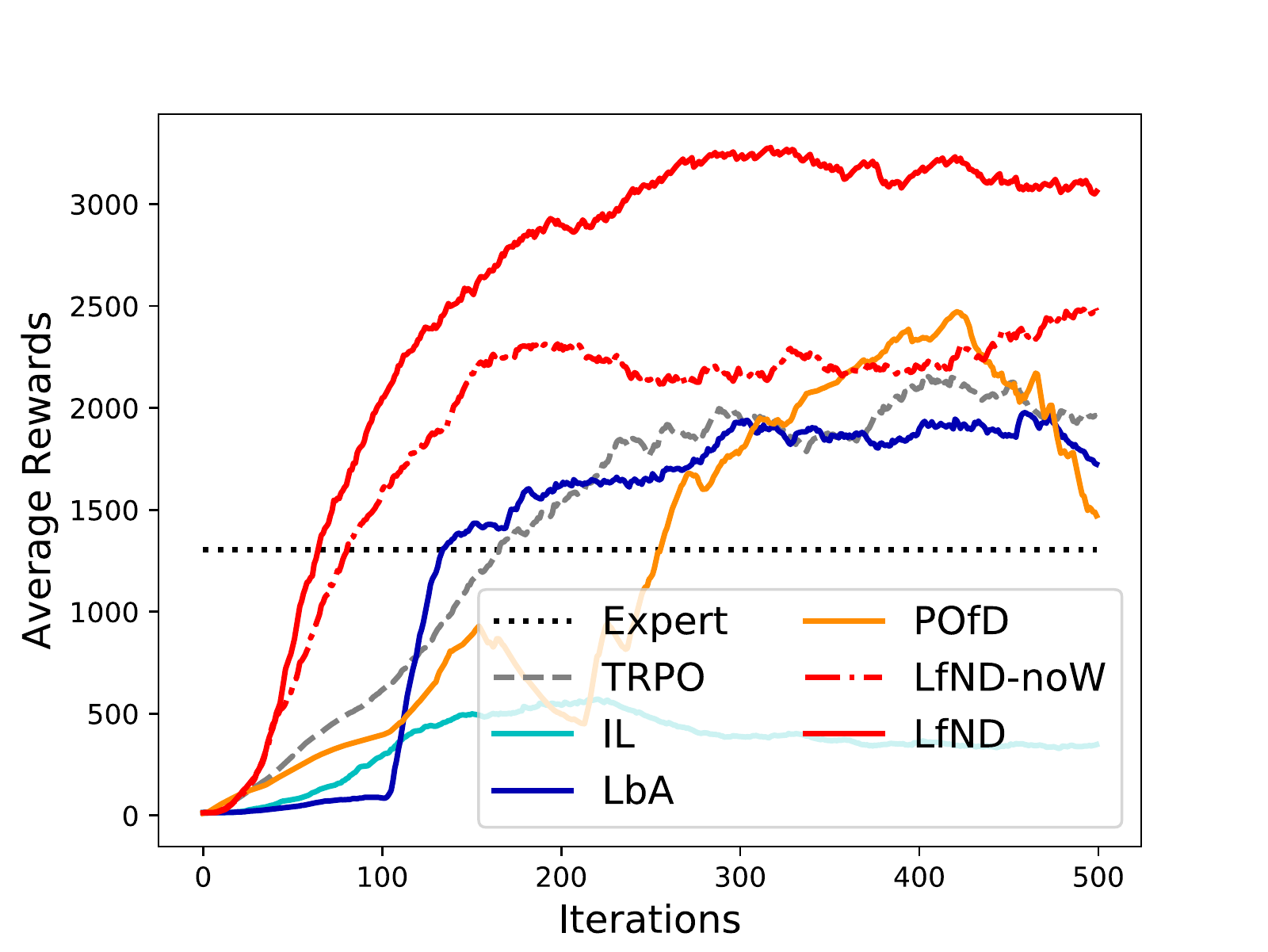}  }
		\subfigure[Humanoid]{
			\label{fig.TRPO-Humanoid}
			\includegraphics[width=0.32\textwidth]{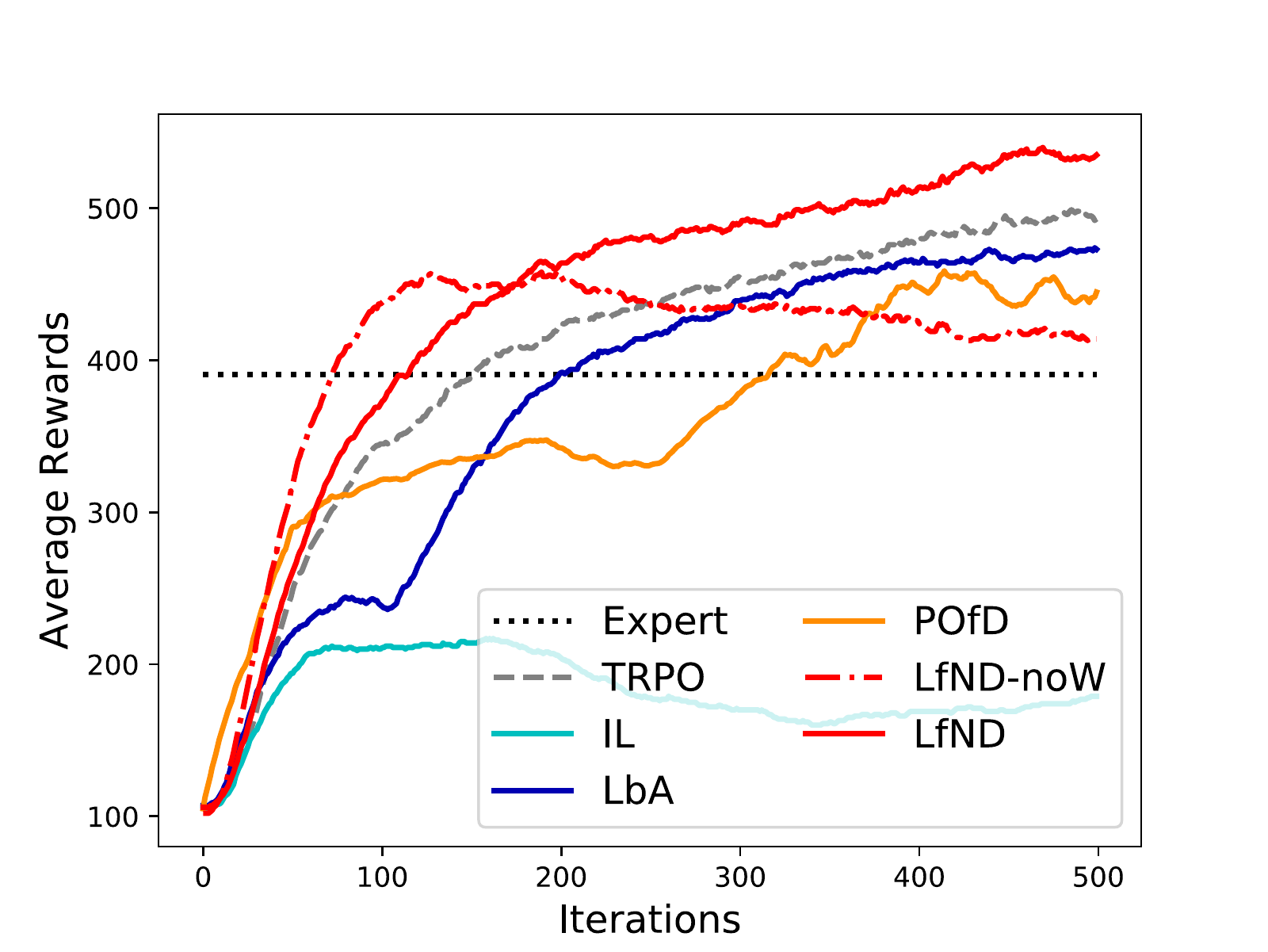}  }
		\subfigure[HumanoidStandup]{
			\label{fig.TRPO-HumanoidStandup}
			\includegraphics[width=0.32\textwidth]{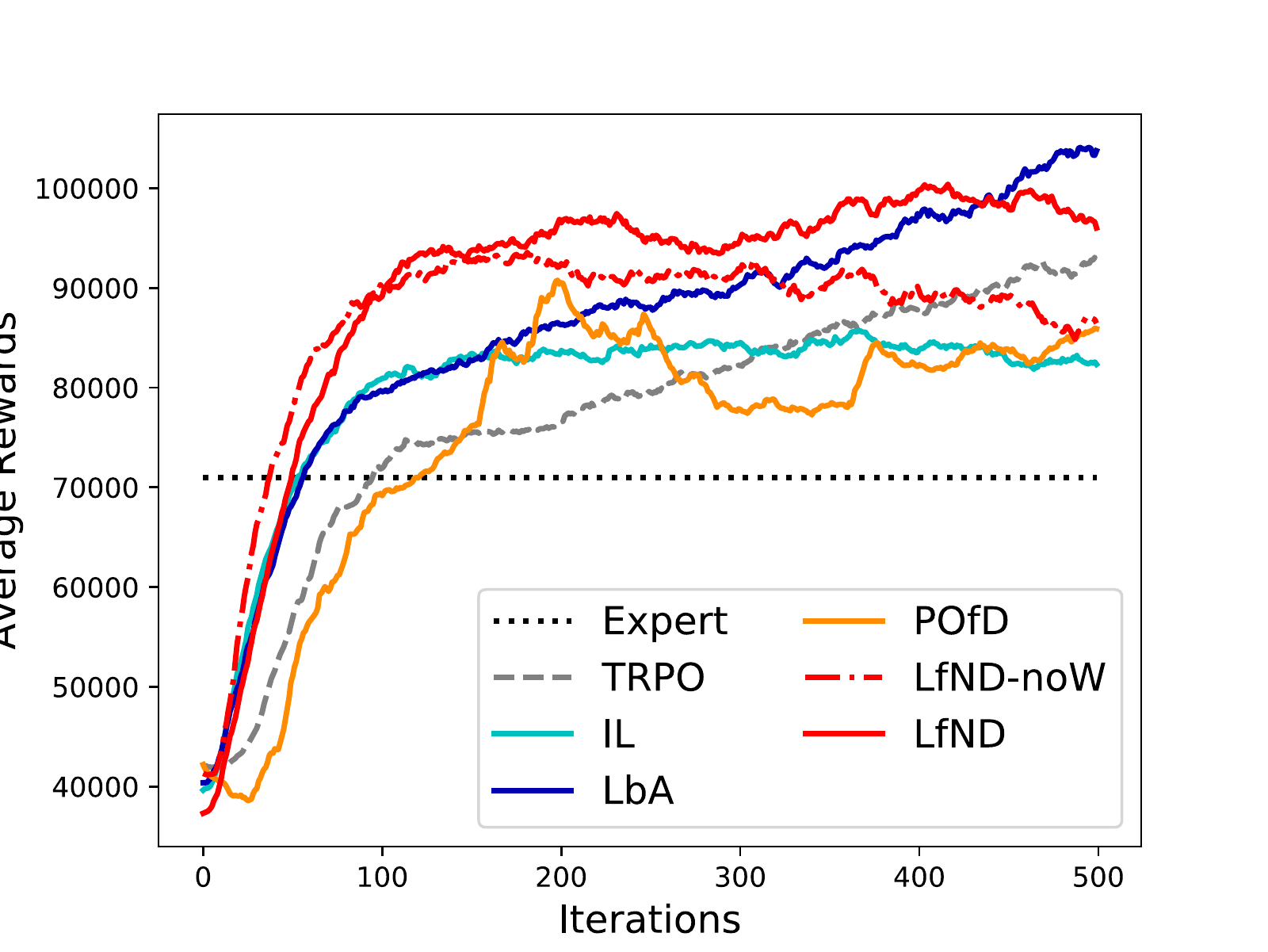}  }
		\subfigure[InvertedPendulum]{
			\label{fig.TRPO-InvertedPendulum}
			\includegraphics[width=0.32\textwidth]{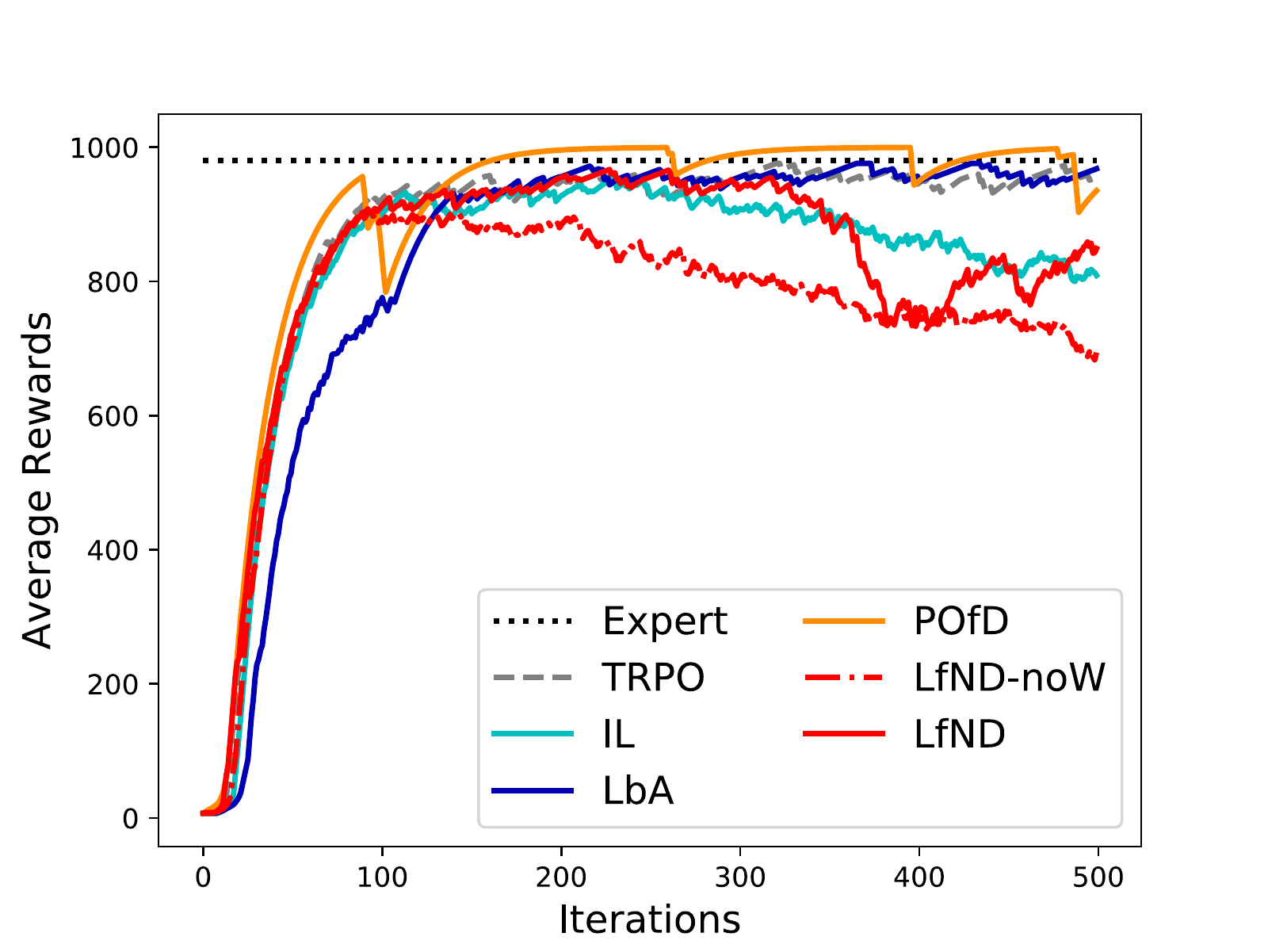}  }
		\subfigure[Walker2d]{
			\label{fig.TRPO-Walker2d}
			\includegraphics[width=0.32\textwidth]{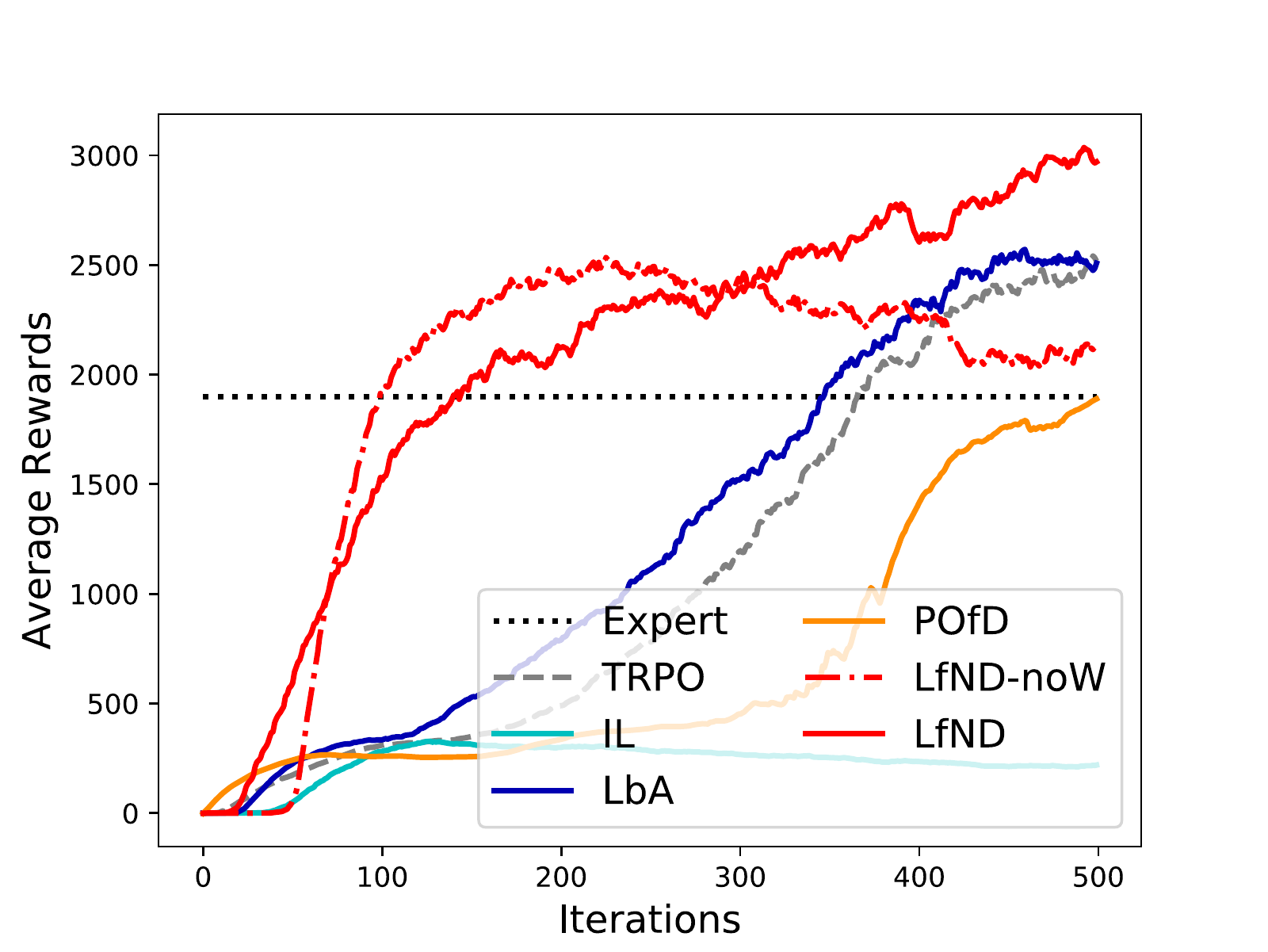}  }				
		\caption{Performance comparison with TRPO as the base RL model.}
		\label{fig.comp.trpo}
	\end{figure*}

	\begin{figure*}[!ht]
		\centering
		\subfigure[HalfCheetah]{
			\label{fig.PPO-HalfCheetah}
			\includegraphics[width=0.32\textwidth]{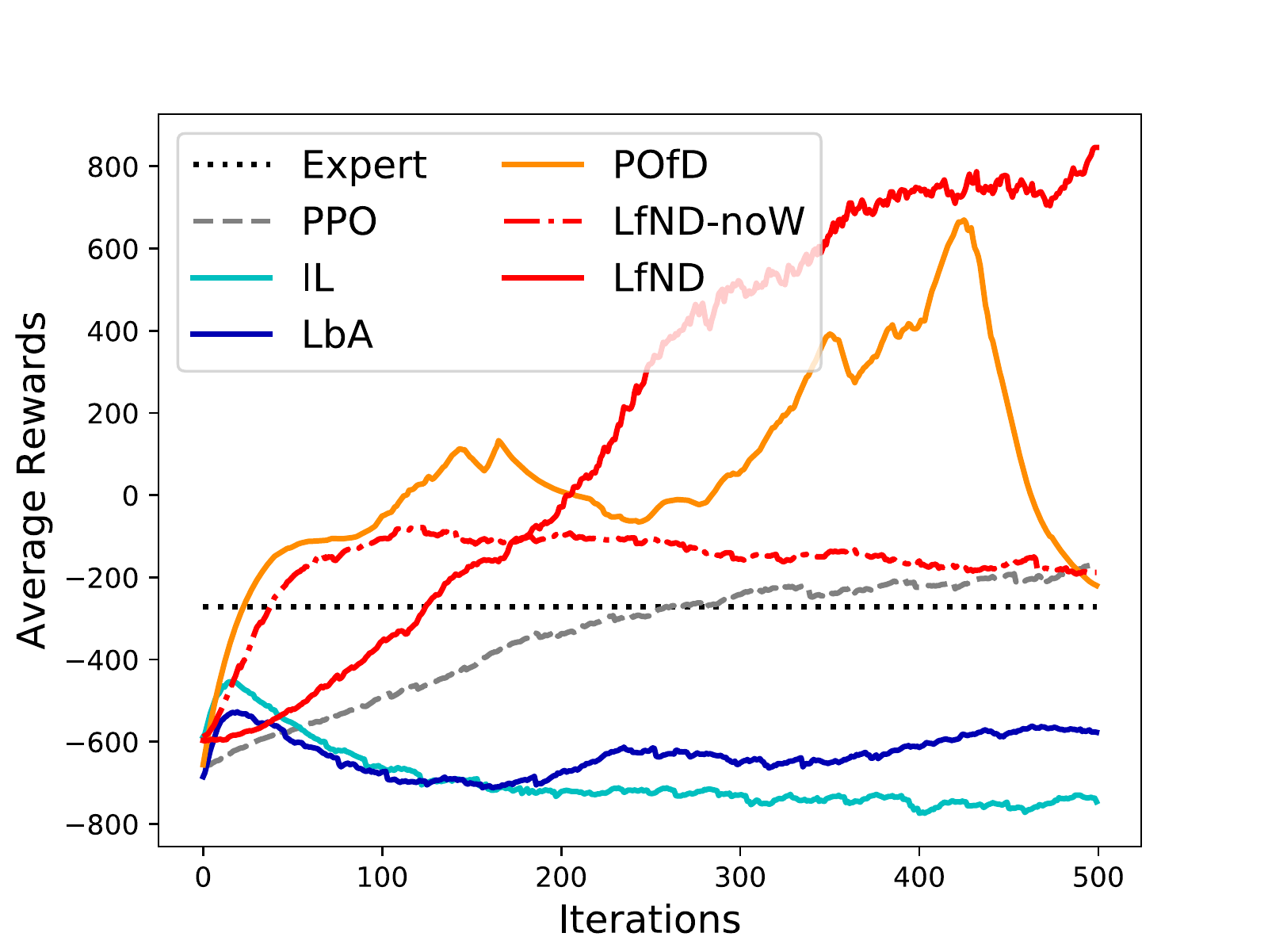}  }
		\subfigure[Hopper]{
			\label{fig.PPO-Hopper}
			\includegraphics[width=0.32\textwidth]{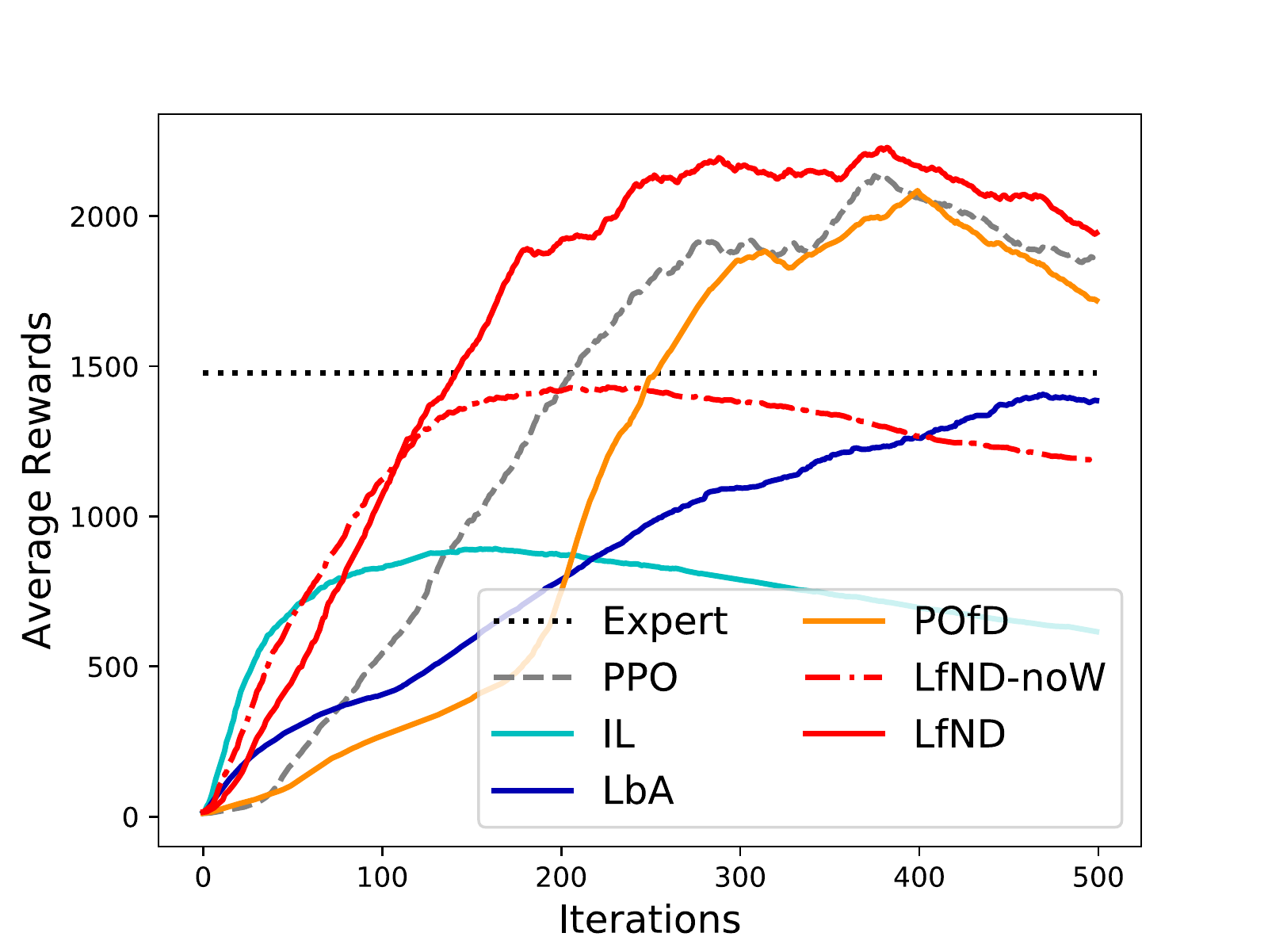}  }
		\subfigure[Humanoid]{
			\label{fig.PPO-Humanoid}
			\includegraphics[width=0.32\textwidth]{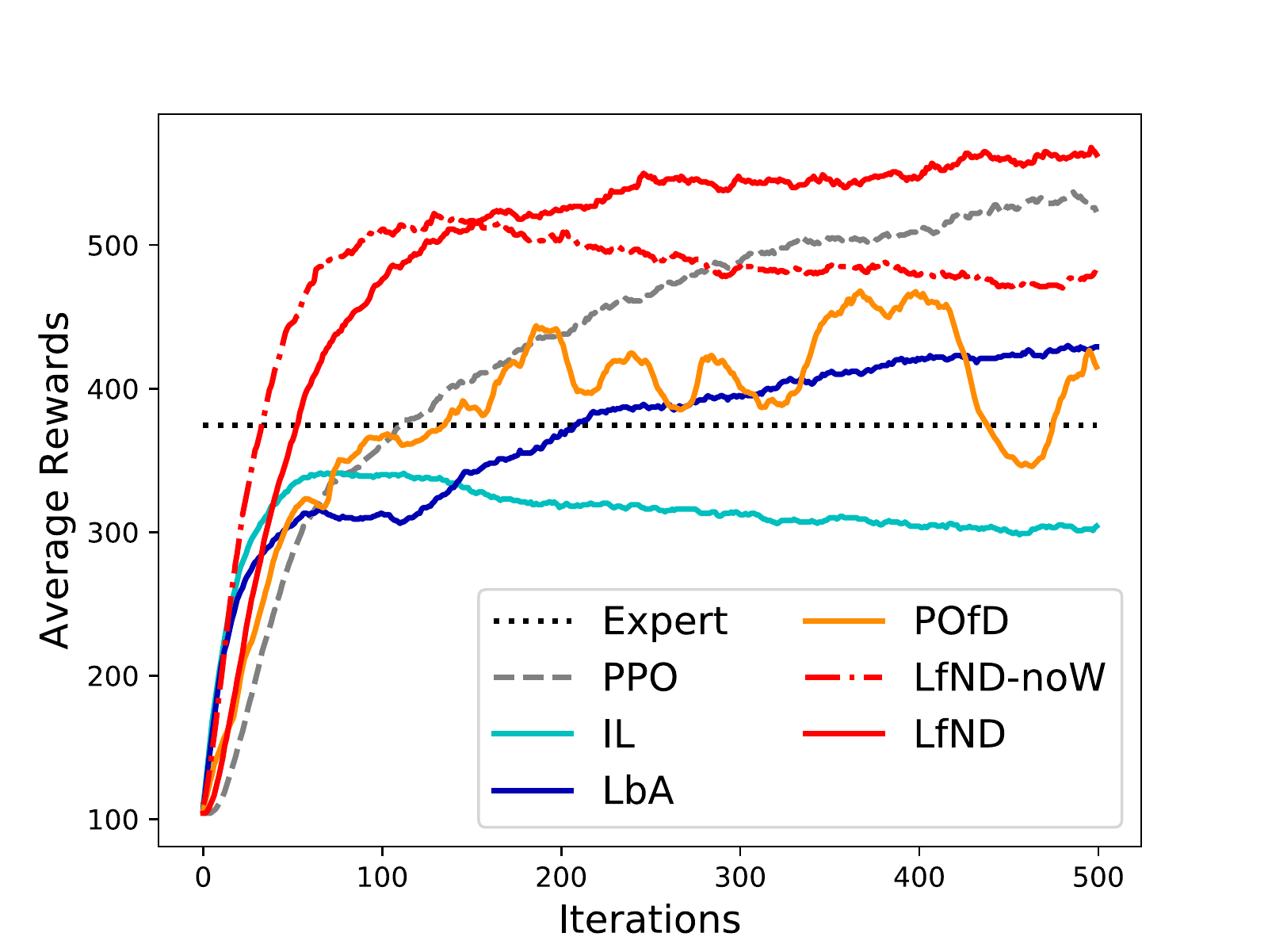}  }
		\subfigure[HumanoidStandup]{
			\label{fig.PPO-HumanoidStandup}
			\includegraphics[width=0.32\textwidth]{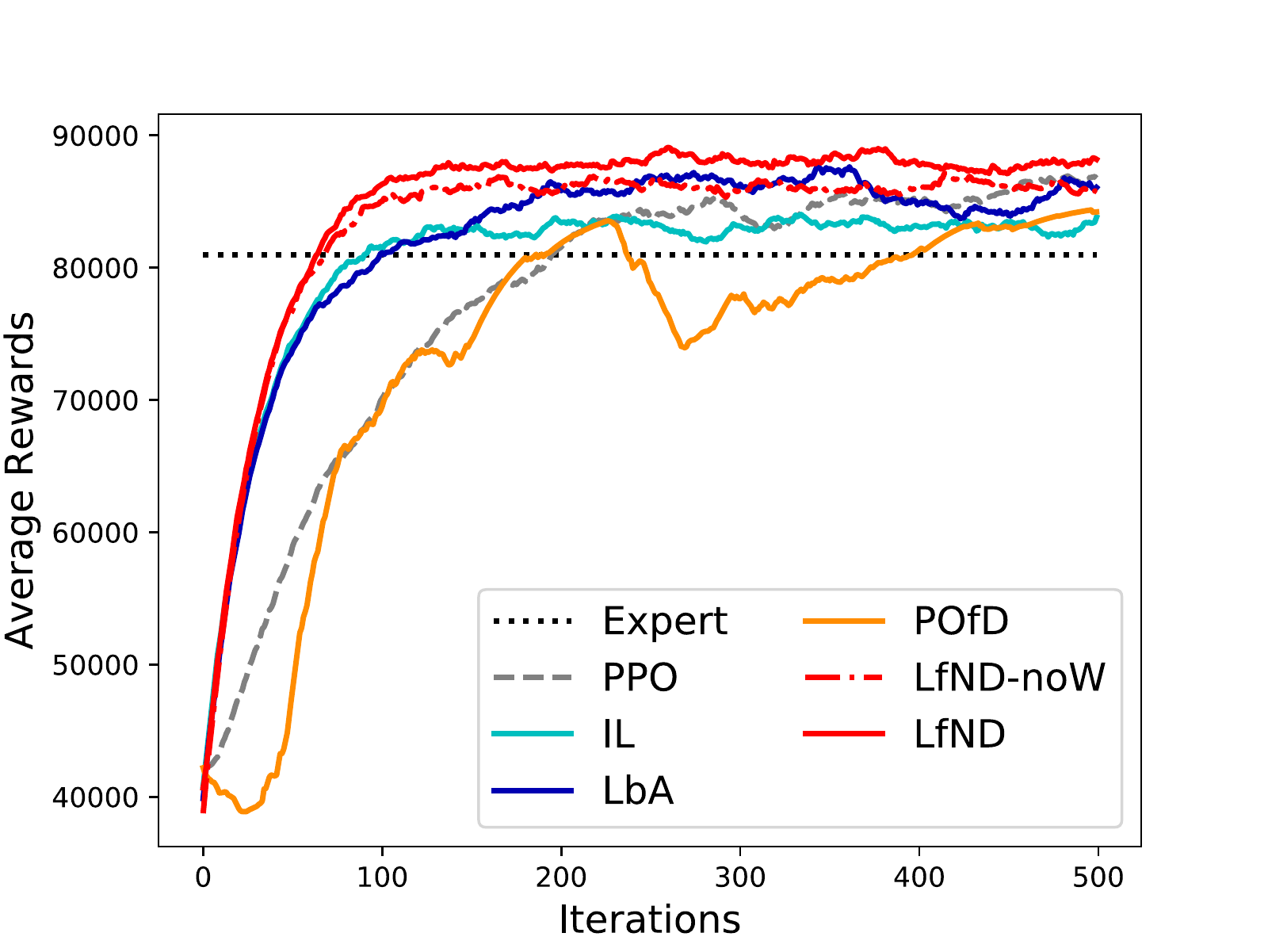}  }
		\subfigure[InvertedPendulum]{
			\label{fig.PPO-InvertedPendulum}
			\includegraphics[width=0.32\textwidth]{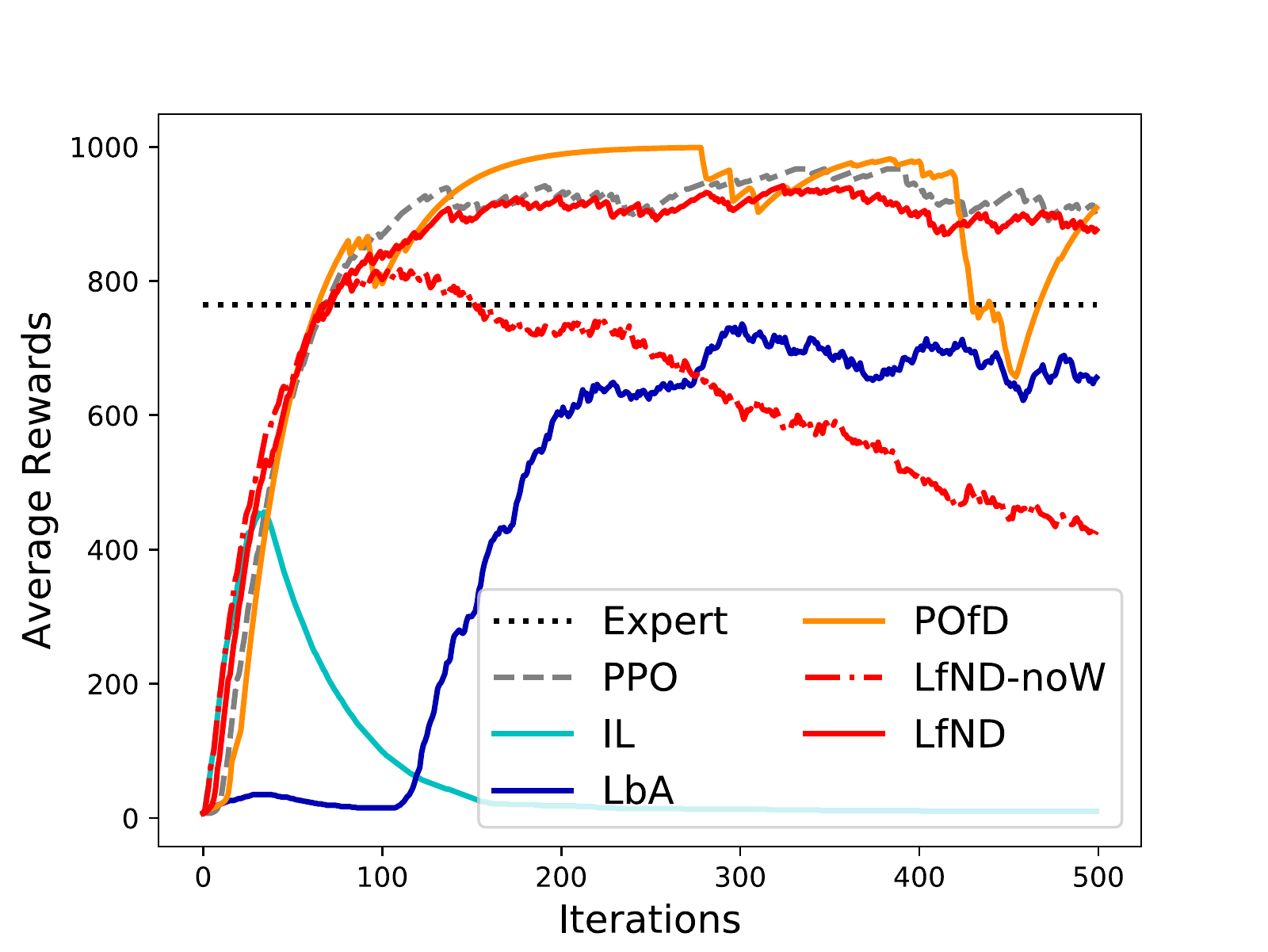}  }
		\subfigure[Walker2d]{
			\label{fig.PPO-Walker2d}
			\includegraphics[width=0.32\textwidth]{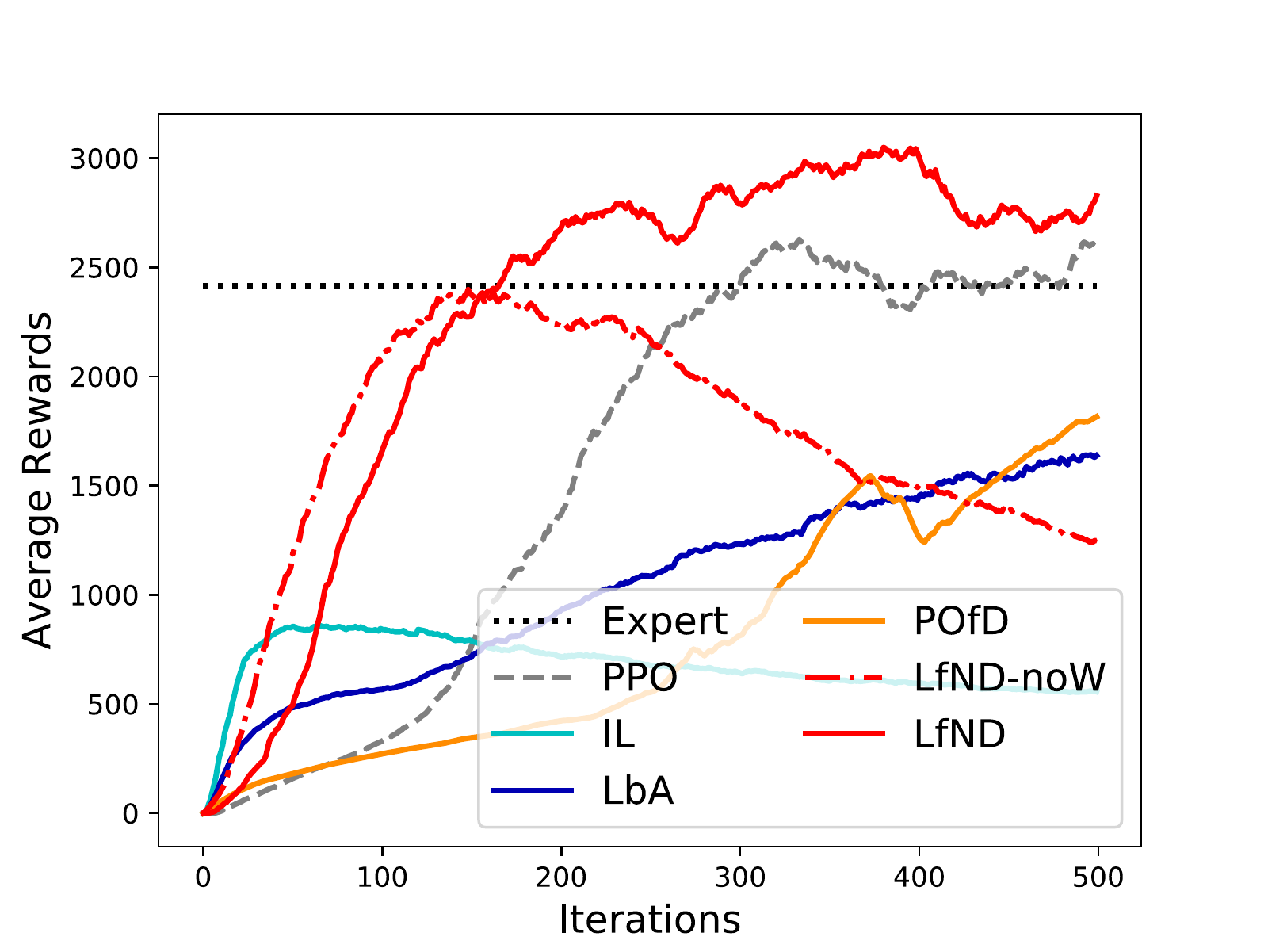}  }
		
		\caption{Performance comparison with PPO as the base RL model.}
		\label{fig.comp.ppo}
	\end{figure*}

	\begin{table*}[!t]
		\newcommand{\tabincell}[2]{\begin{tabular}{@{}#1@{}}#2\end{tabular}}
		\centering
		\caption{The average performance for the compared methods (mean$\pm$std), and the best performance is highlighted in boldface.}
		\label{table:comparisonMean}
		\begin{tabular}{c|c|c c c c c c}
			\toprule
			\hline
			\multirow{2}{*}{ Model } & \multirow{2}{*}{ Methods } & \multicolumn{6}{c}{Environment}\\
			\cline{3-8}
			&&HalfCheetah & Hopper & Humanoid & HumanoidStandup & InvertedPendulum & Walker2d\\
			\hline
			
			\multirow{5}{*}{\tabincell{c}{TRPO}}			
			&\multirow{1}{*}{TRPO}
			& $67.87\pm8.67$ & $1491.85\pm13.61$  & $407.90\pm3.98$ & $78740.67\pm320.28$ & $888.53\pm7.42$ & $1153.15\pm17.25$  \\
			&\multirow{1}{*}{LfND-noW}
			& $\bm{558.99\pm9.49}$ & $1963.89\pm17.19$  & $411.43\pm3.42$ & $88002.65\pm283.67$ & $769.96\pm5.83$ & $2000.67\pm14.72$  \\
			&\multirow{1}{*}{IL}
			& $-450.28\pm1.05$ & $367.98\pm3.52$  & $184.30\pm2.17$ & $80969.97\pm271.42$ & $835.53\pm6.61$ & $224.24\pm2.63$  \\
			&\multirow{1}{*}{LbA}
			& $202.35\pm8.99$ & $1394.61\pm12.49$  & $376.71\pm3.81$ & $87569.64\pm405.12$ & $852.78\pm7.64$ & $1318.84\pm17.31$  \\
			&\multirow{1}{*}{POfD}
			& $103.57\pm9.73$ & $1301.05\pm9.16$  & $362.23\pm2.12$ & $76385.23\pm275.43$ & $\bm{917.03\pm5.86}$ & $717.62\pm11.93$  \\
			&\multirow{1}{*}{LfND}
			& $484.64\pm11.13$ & $\bm{2655.84\pm20.91}$  & $\bm{443.88\pm3.99}$ & $\bm{92033.78\pm373.97}$ & $839.16\pm6.89$ & $\bm{2139.25\pm20.53}$  \\			
			\hline
			\hline
			
			\multirow{5}{*}{\tabincell{c}{PPO}}
			
			&\multirow{1}{*}{PPO}
			& $-338.08\pm1.62$ & $1709.36\pm12.49$  & $436.87\pm4.08$ & $78769.73\pm283.49$ & $\bm{881.34\pm7.30}$ & $1753.05\pm18.47$  \\
			&\multirow{1}{*}{LfND-noW}
			& $-160.35\pm1.03$ & $1411.17\pm8.72$  & $477.72\pm3.94$ & $84537.17\pm297.93$ & $633.41\pm4.38$ & $1797.58\pm9.15$  \\
			&\multirow{1}{*}{IL}
			& $-698.06\pm2.25$ & $877.76\pm5.04$  & $316.46\pm2.64$ & $81577.83\pm269.12$ & $69.63\pm0.55$ & $721.53\pm5.01$  \\
			&\multirow{1}{*}{LbA}
			& $-632.67\pm0.74$ & $1127.36\pm10.48$  & $372.79\pm3.56$ & $83318.53\pm295.48$ & $491.65\pm5.61$ & $1141.59\pm12.09$  \\
			&\multirow{1}{*}{POfD}
			& $71.79\pm2.78$ & $1436.52\pm9.21$  & $384.64\pm1.94$ & $74862.87\pm266.29$ & $866.86\pm5.70$ & $884.03\pm12.13$  \\
			&\multirow{1}{*}{LfND}
			& $\bm{203.15\pm9.04}$ & $\bm{2024.92\pm12.71}$  & $\bm{504.4\pm4.38}$ & $\bm{86100.59\pm300.45}$ & $864.85\pm7.05$ & $\bm{2455.43\pm18.99}$  \\			
			\bottomrule
		\end{tabular}
	\end{table*}

	\begin{table*}[!t]
		\newcommand{\tabincell}[2]{\begin{tabular}{@{}#1@{}}#2\end{tabular}}
		\centering
		\caption{The maximum performance for the compared methods (mean$\pm$std). The best performance is highlighted in boldface.}
		\label{table:comparisonMax}
		\begin{tabular}{c|c|c c c c c c}
			\toprule
			\hline
			\multirow{2}{*}{ Model } & \multirow{2}{*}{ Methods } & \multicolumn{6}{c}{Environment}\\
			\cline{3-8}
			&&HalfCheetah & Hopper & Humanoid & HumanoidStandup & InvertedPendulum & Walker2d\\
			\hline			
			\multirow{5}{*}{\tabincell{c}{TRPO}}
			
			&\multirow{1}{*}{TRPO}
			& $755.80\pm15.02$ & $2245.20\pm28.31$  & $501.20\pm2.87$ & $94034.00\pm414.45$ & $981.80\pm3.27$ & $2614.80\pm32.42$  \\
			&\multirow{1}{*}{LfND-noW}
			& $794.00\pm16.33$ & $2556.80\pm20.89$  & $464.60\pm4.17$ & $94133.60\pm647.27$ & $933.20\pm10.21$ & $2595.20\pm22.12$  \\
			&\multirow{1}{*}{IL}
			& $-240.00\pm13.65$ & $569.20\pm12.52$  & $220.60\pm5.91$ & $86425.40\pm354.36$ & $962.40\pm4.84$ & $365.80\pm7.24$  \\
			&\multirow{1}{*}{LbA}
			& $820.20\pm16.43$ & $2038.00\pm38.51$  & $478.60\pm2.58$ & $\bm{105440.40\pm578.23}$ & $981.40\pm2.87$ & $2636.00\pm31.29$  \\
			&\multirow{1}{*}{POfD}
			& $918.80\pm19.54$ & $2549.20\pm25.99$  & $462.80\pm3.12$ & $94763.20\pm1375.20$ & $\bm{1000.00\pm0.02}$ & $1943.20\pm15.83$  \\
			&\multirow{1}{*}{LfND}
			& $\bm{1017.80\pm16.46}$ & $\bm{3317.20\pm28.56}$  & $\bm{545.00\pm3.43}$ & $101466.40\pm486.12$ & $976.00\pm4.12$ & $\bm{3106.40\pm27.44}$  \\			
			\hline
			\hline
			
			\multirow{5}{*}{\tabincell{c}{PPO}}
			
			&\multirow{1}{*}{PPO}
			& $-165.40\pm2.71$ & $2640.20\pm66.91$  & $541.80\pm2.99$ & $87324.20\pm243.71$ & $994.80\pm2.87$ & $2875.60\pm54.50$  \\
			&\multirow{1}{*}{LfND-noW}
			& $-72.40\pm6.83$ & $1954.00\pm17.08$  & $528.60\pm5.39$ & $87822.40\pm415.35$ & $876.60\pm15.87$ & $2705.20\pm31.56$  \\
			&\multirow{1}{*}{IL}
			& $-434.20\pm21.25$ & $1664.80\pm126.93$  & $351.20\pm5.51$ & $84479.20\pm373.56$ & $592.60\pm56.83$ & $1362.40\pm153.83$  \\
			&\multirow{1}{*}{LbA}
			& $-508.60\pm22.31$ & $1629.40\pm26.94$  & $435.40\pm2.42$ & $88370.80\pm440.87$ & $774.00\pm15.80$ & $1722.60\pm25.19$  \\
			&\multirow{1}{*}{POfD}
			& $706.40\pm34.08$ & $2524.20\pm36.01$  & $474.60\pm4.17$ & $84709.80\pm74.62$ & $\bm{999.80\pm0.01}$ & $1968.80\pm22.83$  \\
			&\multirow{1}{*}{LfND}
			& $\bm{865.80\pm21.59}$ & $\bm{2815.20\pm47.90}$  & $\bm{571.00\pm3.58}$ & $\bm{89660.20\pm262.49}$ & $962.20\pm6.18$ & $\bm{3174.40\pm38.77}$  \\			
			\bottomrule
		\end{tabular}
	\end{table*}

\subsection{Performance comparison}
	We evaluate the performance of compared methods by plotting the reward curve with the number of training iterations increases. The results with TRPO as the base model are presented in Figure \ref{fig.comp.trpo}, while the results with PPO as the base model are shown in Figure \ref{fig.comp.ppo}. It is worthy to note that when comparing with a specific RL base algorithm, we use the same algorithm to implement LfND and all other methods for fair comparison. In addition, we also show the average reward from noisy demonstrations in the figures, denoted by \textbf{Expert}.
	
	From Figures \ref{fig.comp.trpo} and \ref{fig.comp.ppo}, we can observe that no matter which base model is used, the proposed LfND approach outperforms the other methods in most cases. LfND can achieve higher reward with fewer training iterations in general. The IL method which imitates the demonstration without exploring environment is not effective in all environments, and typically cannot reach the average reward of demonstrations. This phenomenon implies that it is important to learn the policy by both exploiting the demonstration and exploring the environment, especially when the demonstrations are noisy. When comparing with the methods that only exploring the environment, i.e., TRPO or PPO, our method is always superior by utilizing the supervised information from demonstration. In contrast, the other two methods POfD and LbA, which also combine environment exploring and demonstration exploiting, are less robust. They outperforms TRPO or PPO in some environments but loss in the others, probably misled by the noisy demonstrations since they utilize all trajectories without distinction. The LfND-noW approach, which is a variant of the proposed approach without weighting scheme, can improve the performance quickly at the early stage but fail to achieve higher performance persistently. One possible reason is that at the early stage, the learned policy is quite poor, and thus most demonstrations are superior than it and can provide useful information to guide the training. After some iterations, when the policy getting more effective, the noisy demonstrations will hurt the performance. This also validates that a specific demonstration may contribute differently at different learning stages, and thus it is important to adaptively adjust the weights of demonstrations as in LfND. The results in Figures \ref{fig.comp.trpo} and \ref{fig.comp.ppo} are consistent in general, validating that the proposed strategy can be effectively incorporated with different reinforcement learning base models.
	
	\begin{figure*}[!ht]
		\centering
		\subfigure[HalfCheetah]{
			\includegraphics[width=0.32\textwidth]{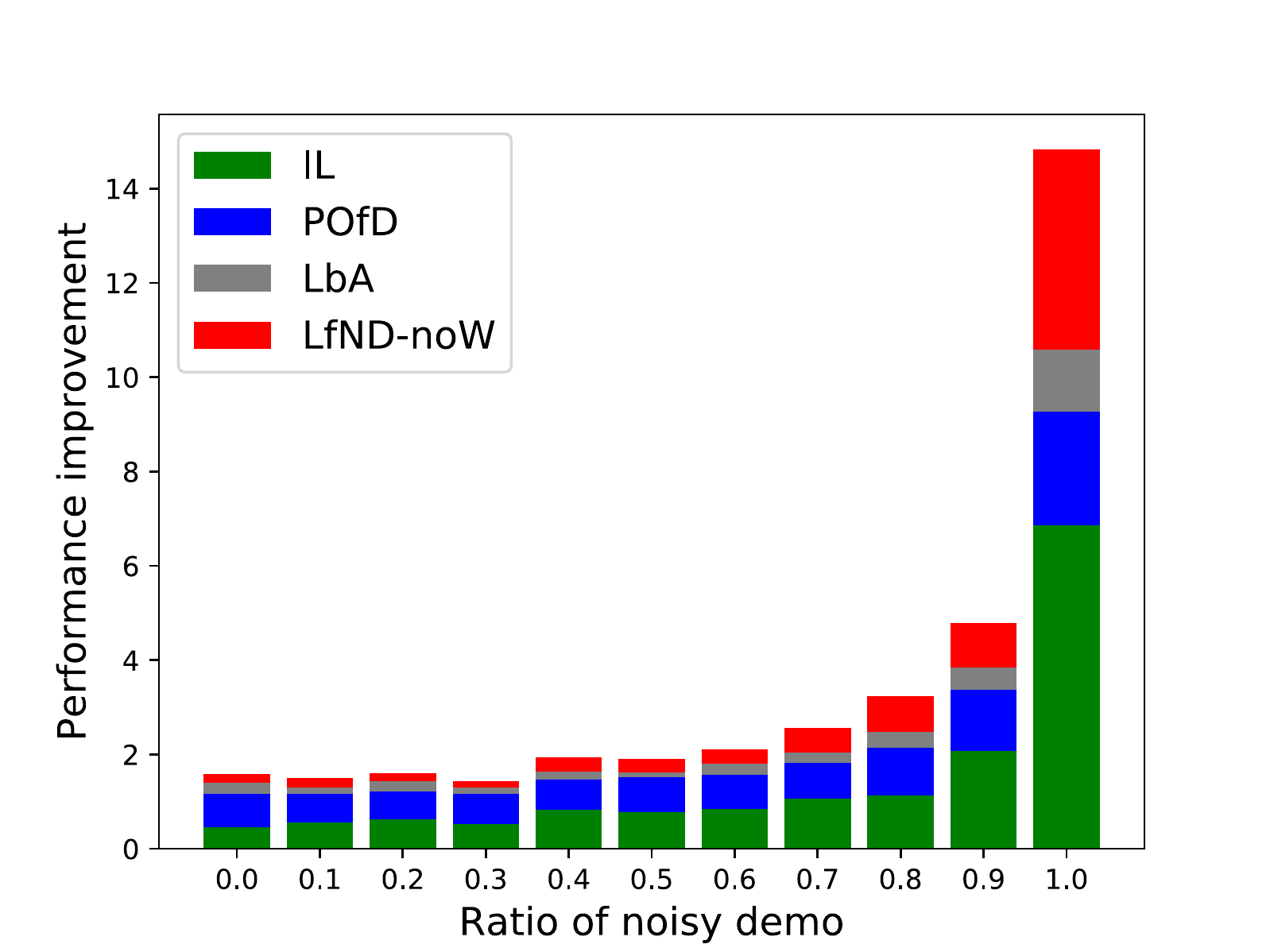}  }
		\subfigure[Humanoid]{
			\includegraphics[width=0.32\textwidth]{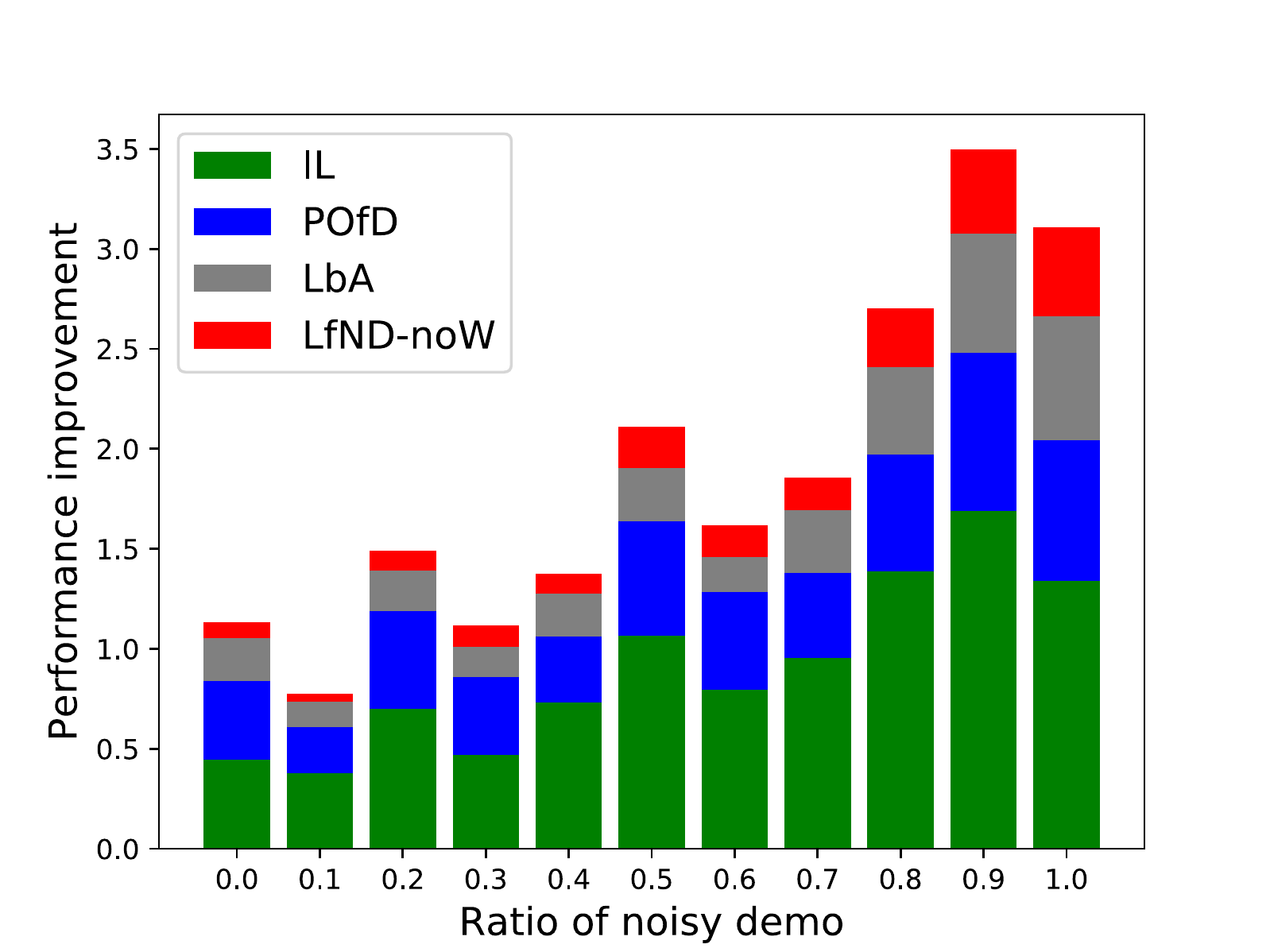}  }
		\subfigure[Walker2d]{
			\includegraphics[width=0.32\textwidth]{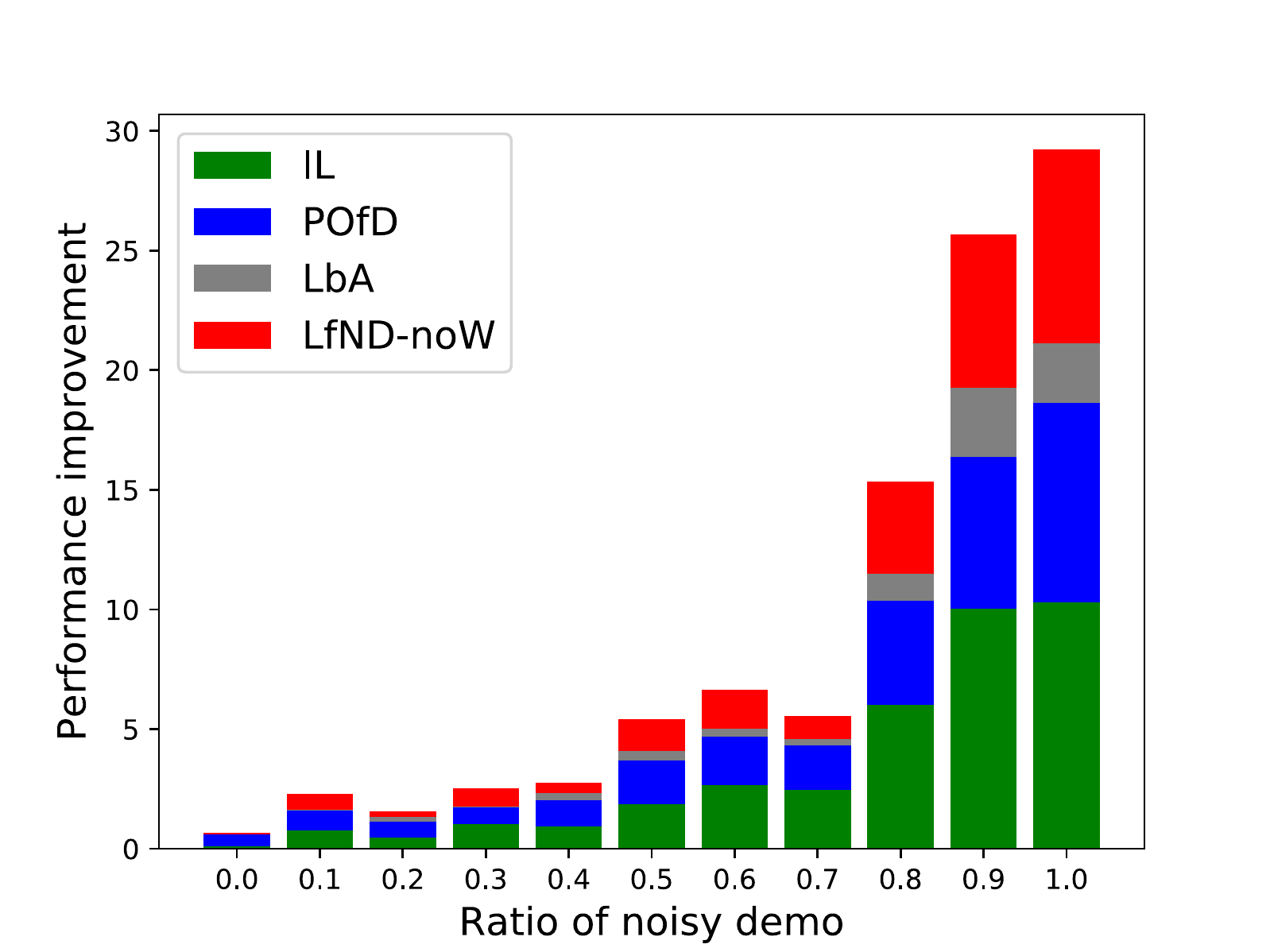}  }
		
		\caption{Performance comparison with different noise ratios of demonstrations.}
		\label{fig.performanceNoisy}
	\end{figure*}

	\begin{figure*}[!ht]
		\centering
		\subfigure[HalfCheetah]{
			\includegraphics[width=0.32\textwidth]{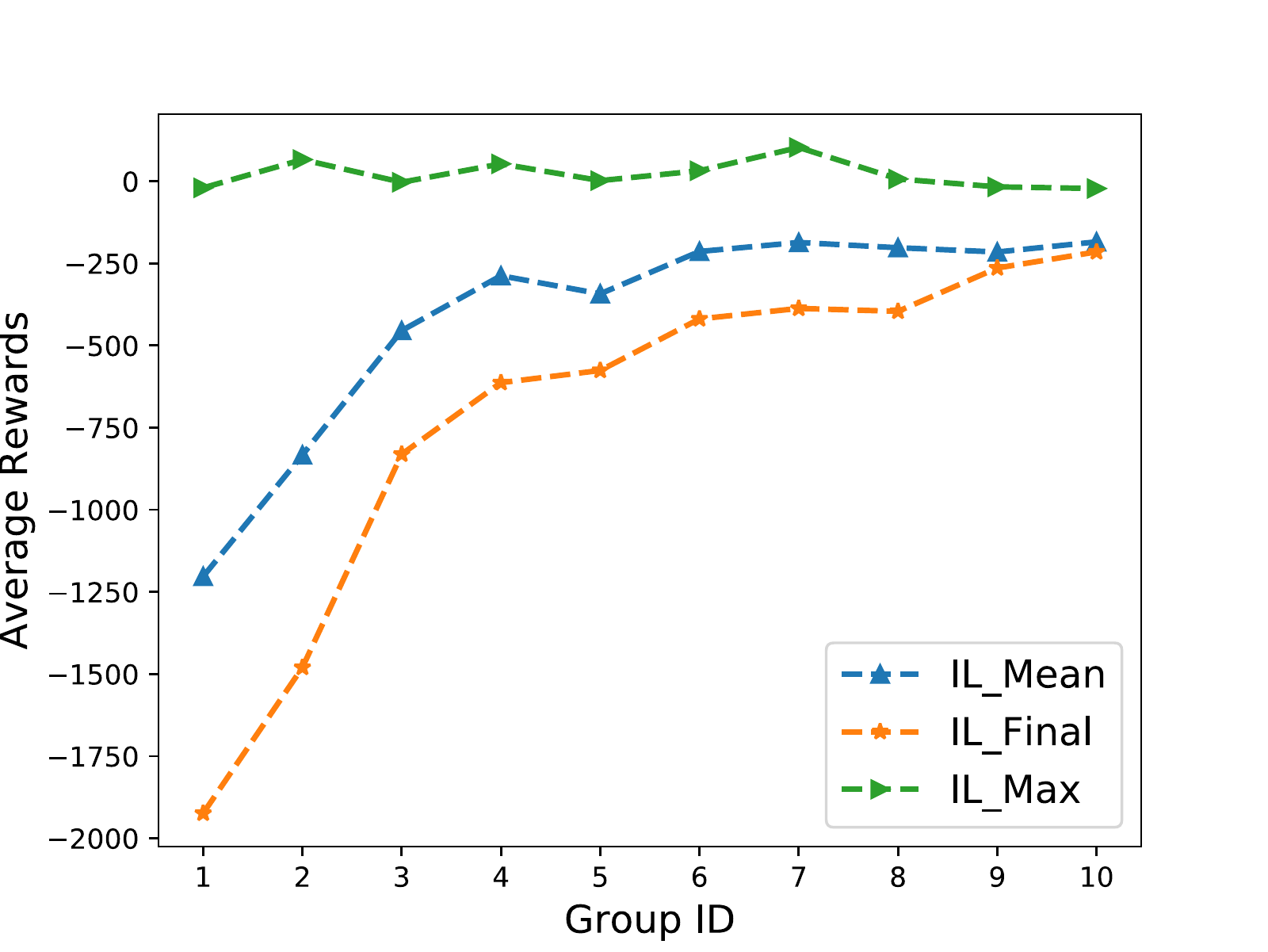}  }
		\subfigure[Humanoid]{
			\includegraphics[width=0.32\textwidth]{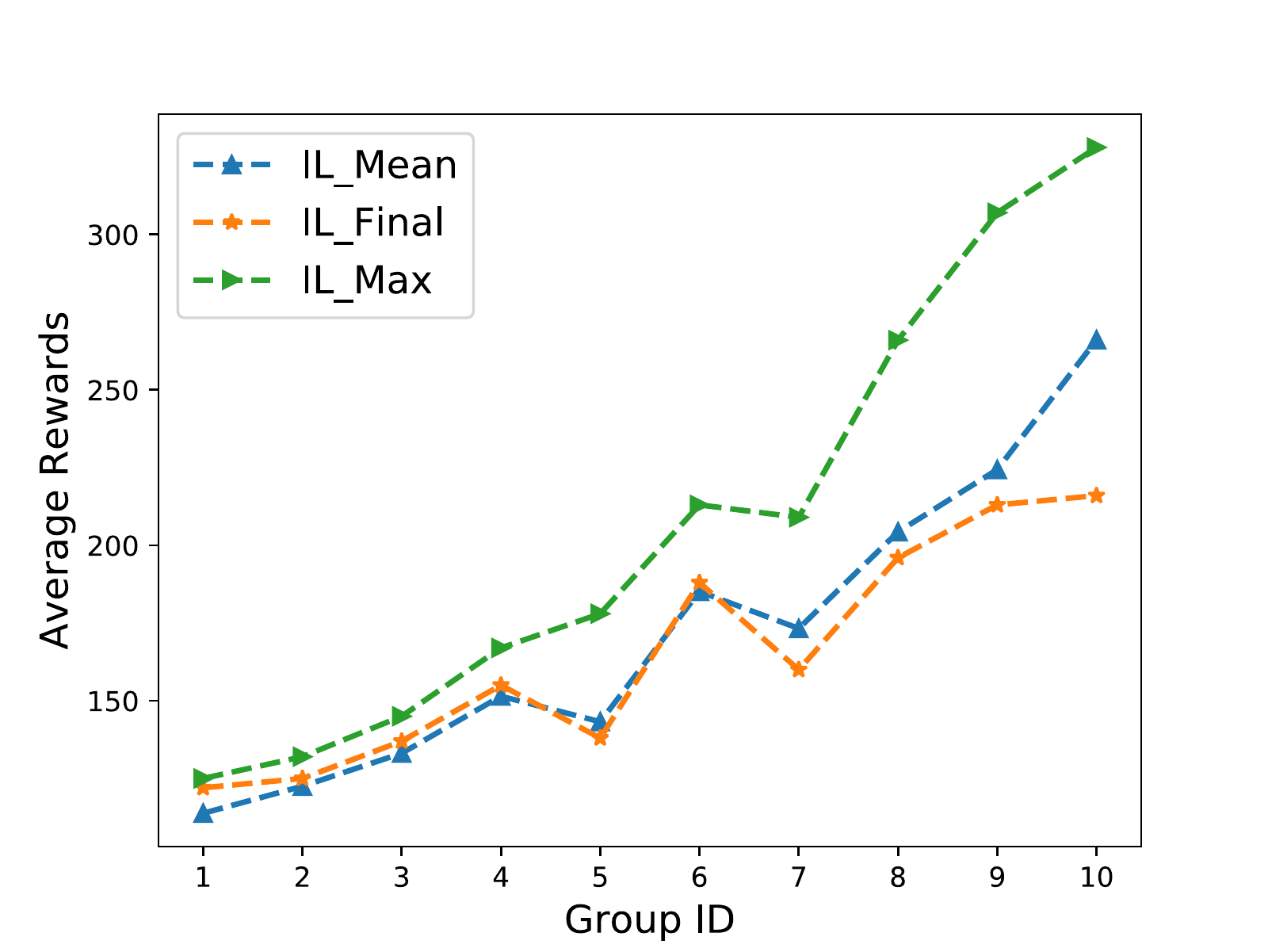}  }
		\subfigure[Walker2d]{
			\includegraphics[width=0.32\textwidth]{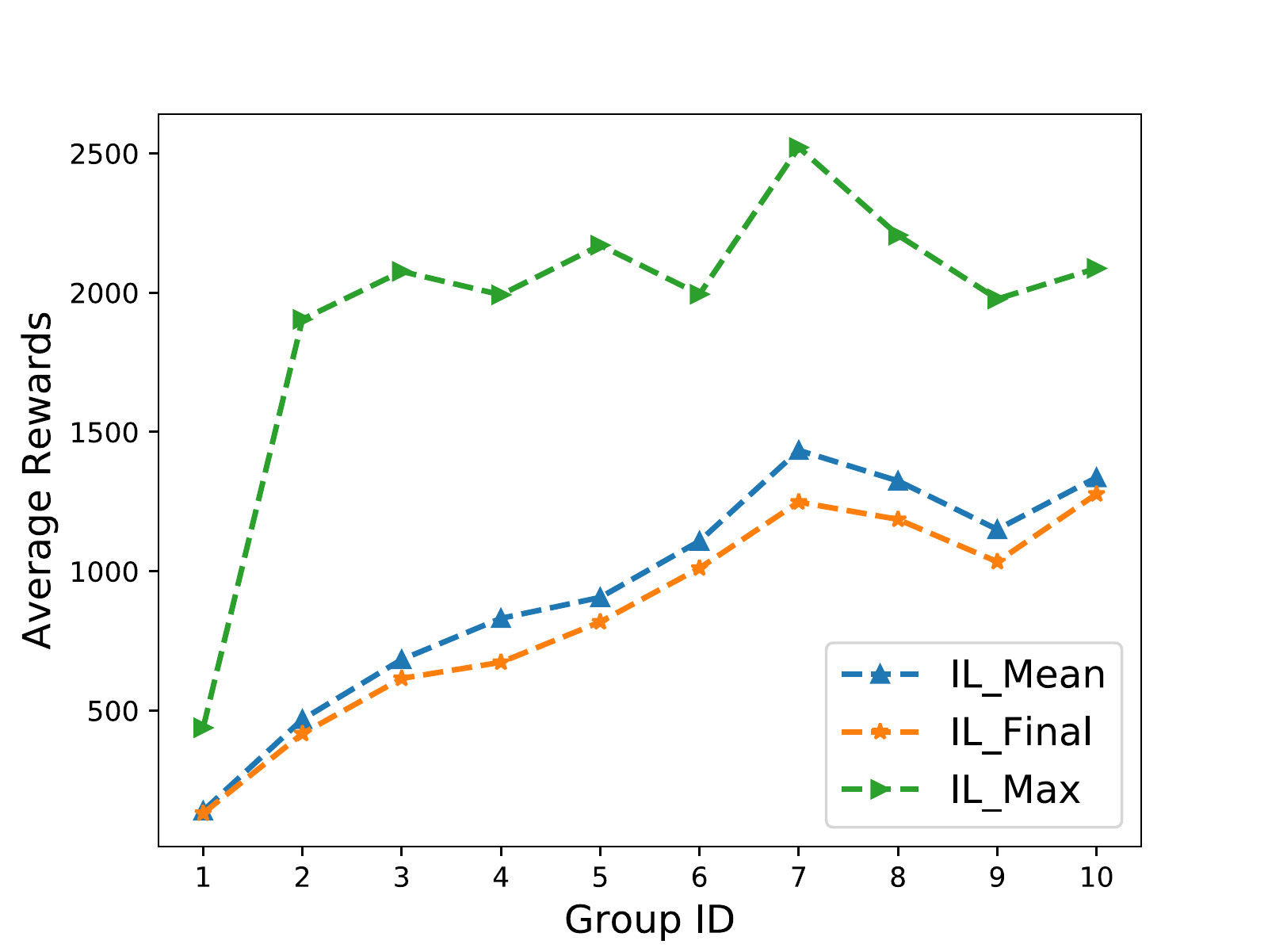}  }
		
		\caption{Performances of IL with different demonstrations. Demonstrations are categorized into 10 groups accordding to their weights. A smaller group ID implies smaller weights for the demonstrations in the group.}
		\label{fig.performanceIL}
	\end{figure*}
	
	To further validate the superiority of our method, we also show the average and maximum reward achieved by different algorithms within 500 learning iterations. The mean results over 5 times repeated experiments with different random seeds are recorded. Results on average reward and maximum reward are reported in Tables~\ref{table:comparisonMean} and ~\ref{table:comparisonMax}, respectively. It can be observed that the proposed approach LfND outperforms the other methods in most cases with regard to both average reward and maximum reward. In other words, the LfND method can consistently keep a higher performance in the learning process, and also obtain a better final policy.

	\begin{figure}[!h]
		\centering
		\subfigure[TRPO-HalfCheetah]{
			\includegraphics[width=0.22\textwidth]{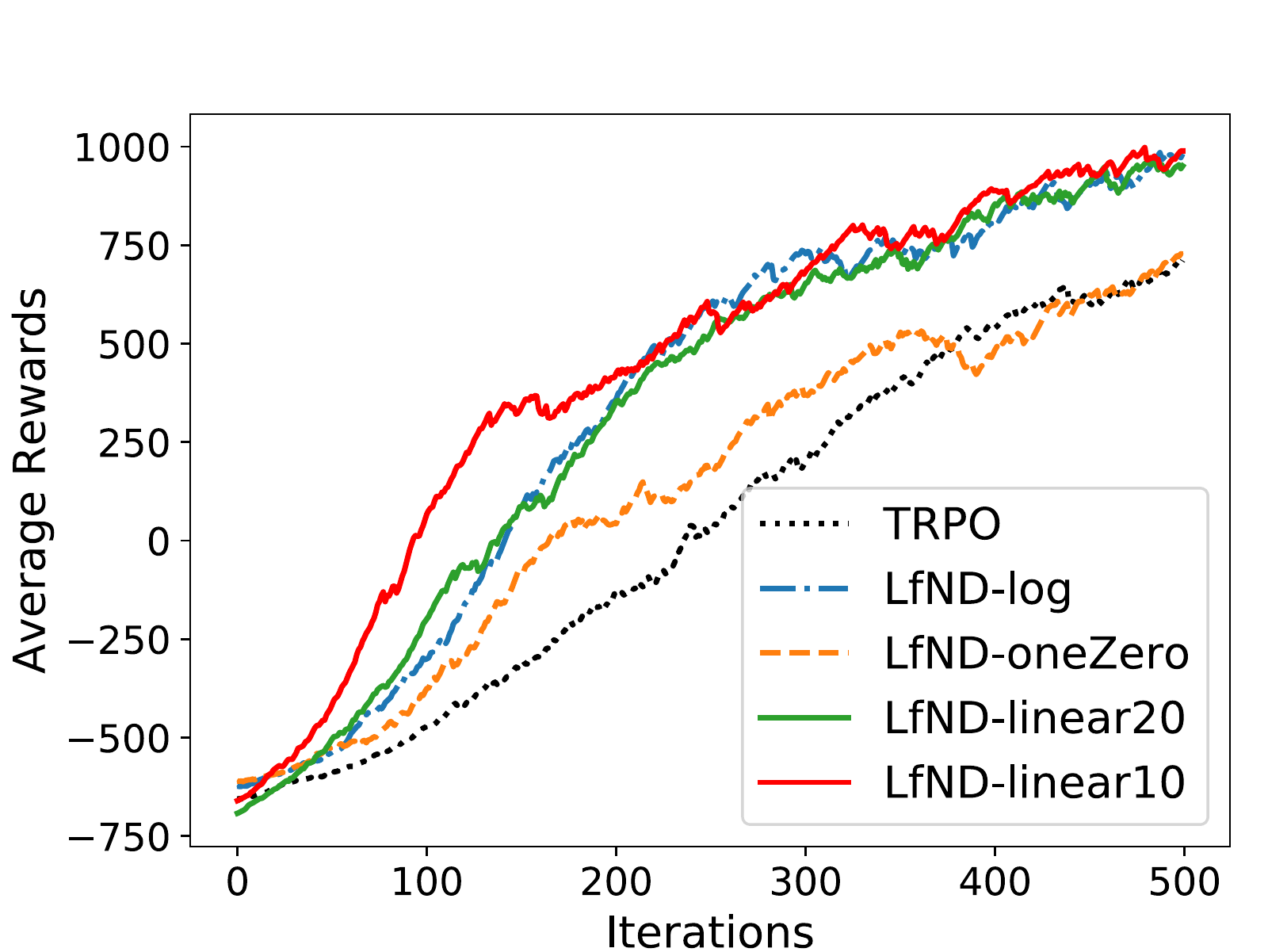}  }
		\subfigure[PPO-HalfCheetah]{
			\includegraphics[width=0.22\textwidth]{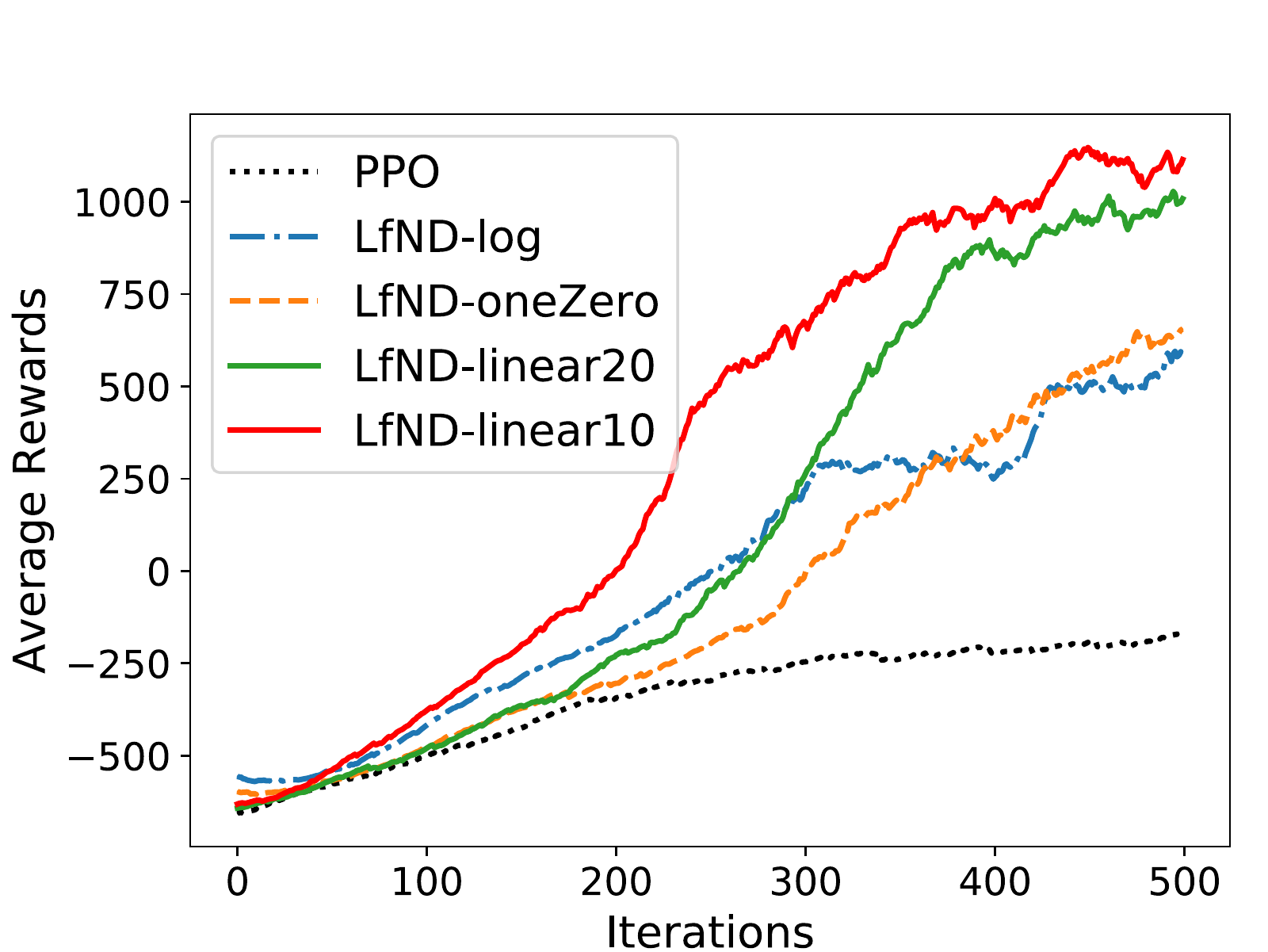}  }
		\subfigure[TRPO-Humanoid]{
			\includegraphics[width=0.22\textwidth]{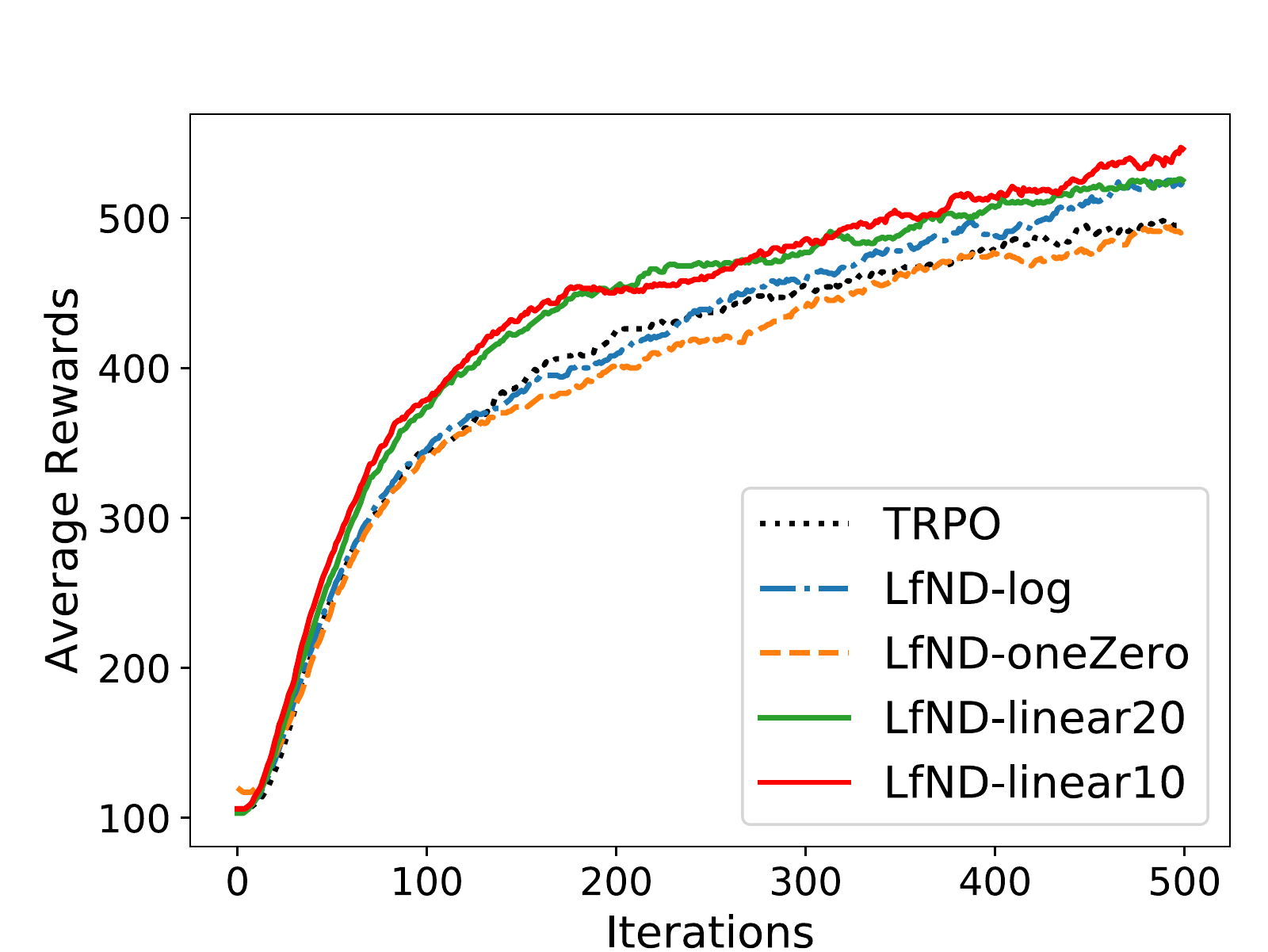}  }
		\subfigure[PPO-Humanoid]{
			\includegraphics[width=0.22\textwidth]{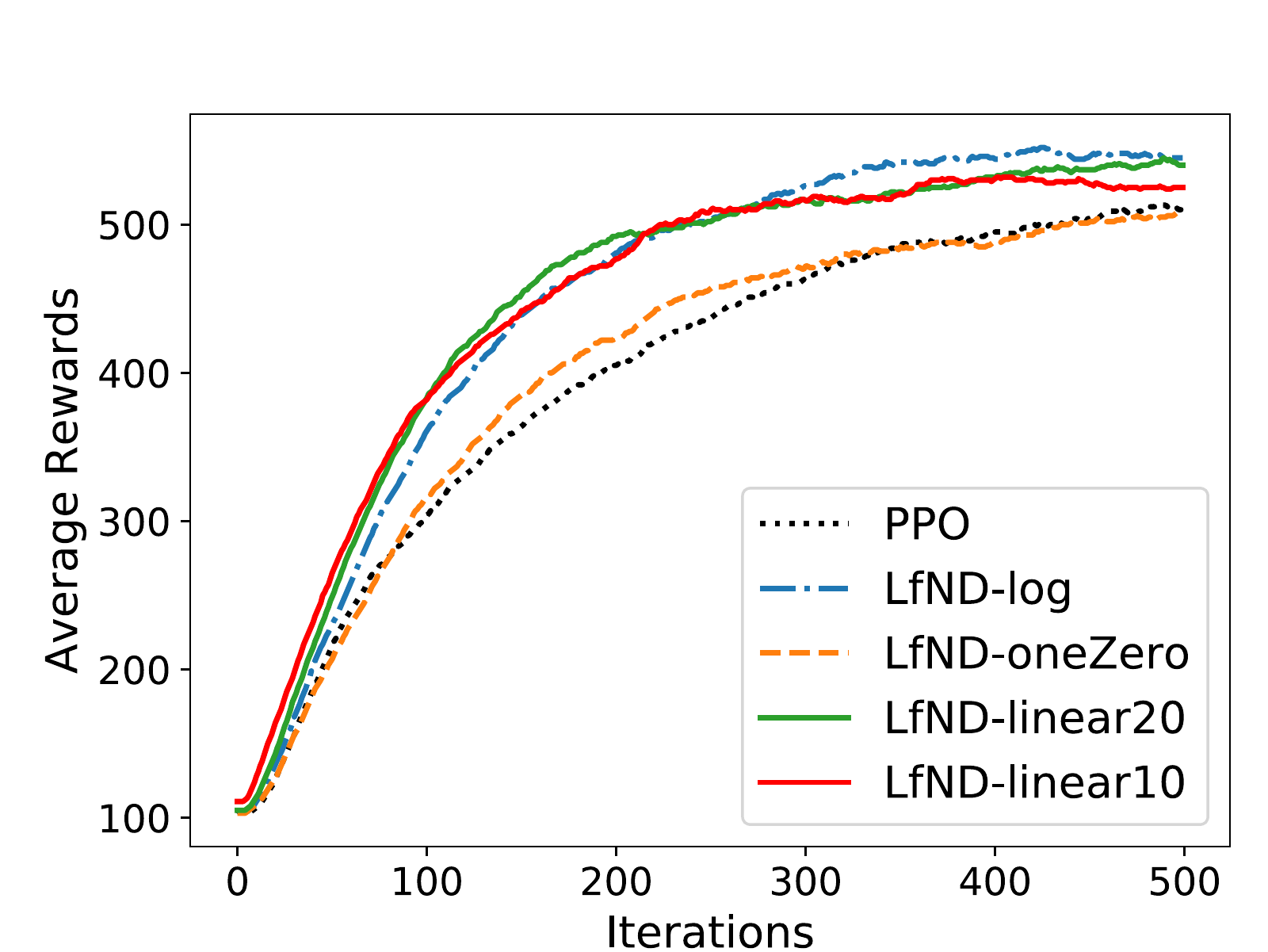}  }
		\subfigure[TRPO-Walker2d]{
			\includegraphics[width=0.22\textwidth]{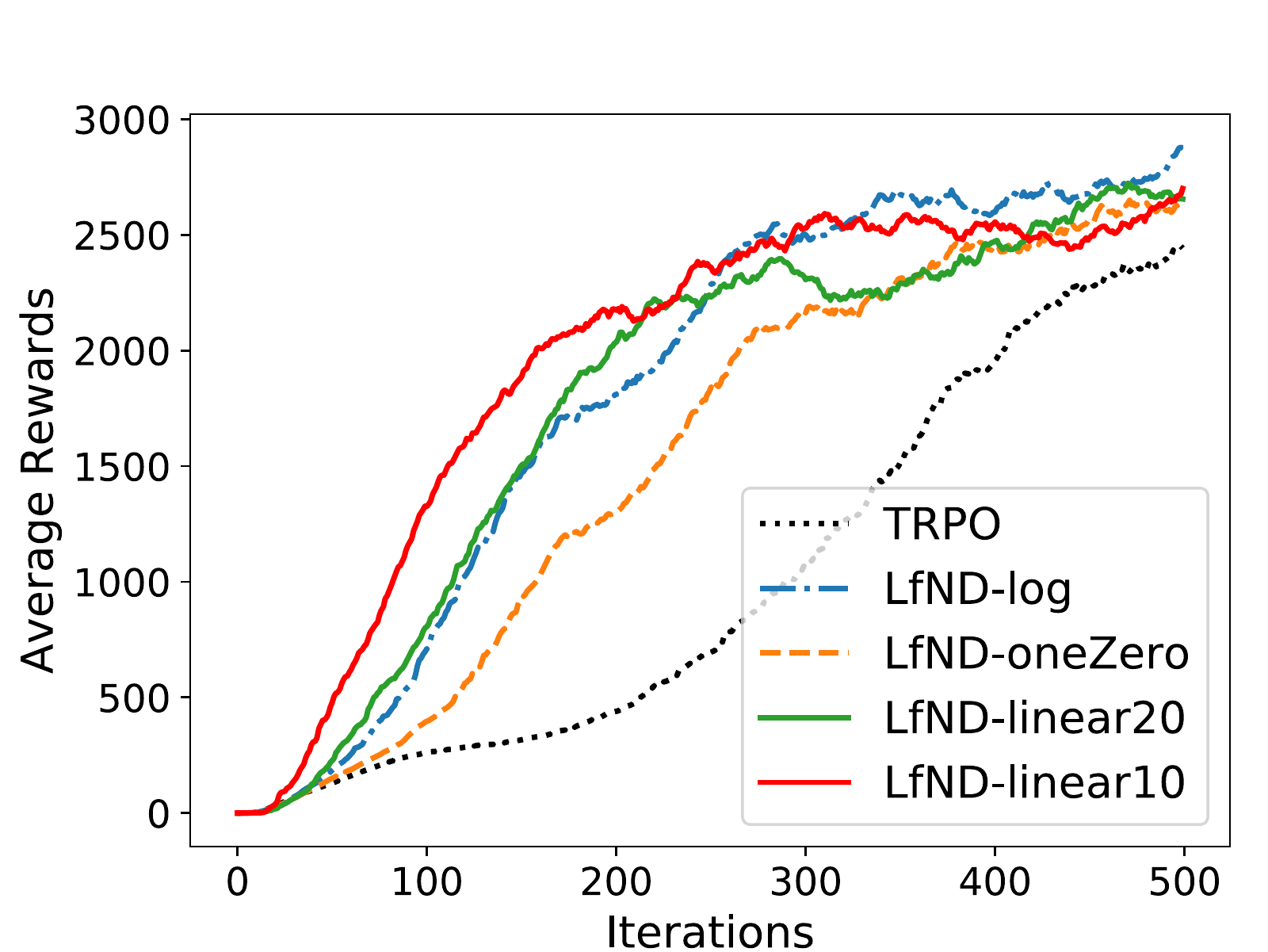}  }
		\subfigure[PPO-Walker2d]{
			\includegraphics[width=0.22\textwidth]{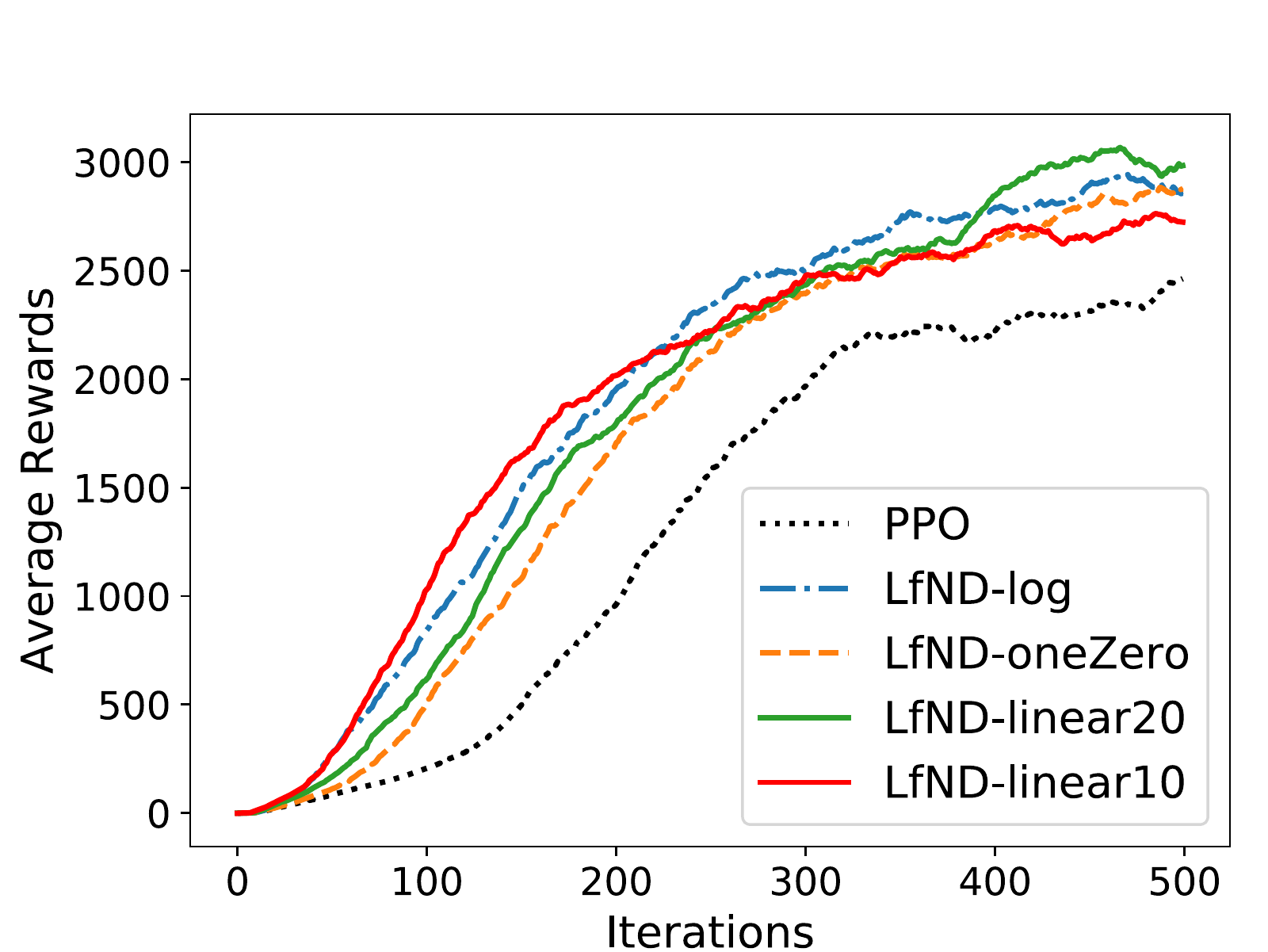}  }
		
		\caption{Performance comparison with different weight forms.}
		\label{fig.performanceComparison2}
	\end{figure}

	\subsection{Study on different noise ratios}
	In this subsection, to further examine the robustness of the proposed method, we perform the experiments with different noise ratios of demonstrations. Specifically, we vary the noise ratio from 0 to 1 with interval of 0.1, where 0 ratio means there is no noise demonstration. For each ratio, we perform the experiments for five methods and record their average reward over 500 learning iterations. Then we calculate the improvement ratio of the proposed method over the other methods. Figure \ref{fig.performanceNoisy} shows the overall improvements of our method over other methods, one color for a method. Due to space limitation, we report the results with TRPO as the base model in three environments, i.e., HalfCheetah, Humanoid and Walker2d. It can be observed that the LfND approach always outperforms other compared methods with all ratios of noise demonstrations. With the increase of noise ratio, the superiority of LfND over the other methods becomes more significant. These results indicate that the proposed LfND method can effectively and robustly learn the policy from noisy demonstrations.
	
	\subsection{Examination on noise identification}
	In this subsection, we further perform experiments to examine whether potential utility of demonstrations are accurately estimated by our algorithm. Specifically, we firstly calculate the average weight of each demonstration instance during the learning process. Then we sort all the instances in increasing order of their average weights, and separate them into 10 groups with equal size. A smaller group ID implies smaller weights for the demonstrations in the group. After that, we fed the demonstrations from each group into the imitation learning (IL) algorithm to learn a policy, and record its performance on average reward, maximum reward and final reward during 500 iterations, respectively.
	
	In Figure~\ref{fig.performanceIL}, the reward curves with different groups are plotted in HalfCheetah, Humanoid and Walker2d. The \textbf{IL\_Mean}, the \textbf{IL\_Final} and the \textbf{IL\_Max} denote the mean, final, maximum rewards respectively in 10 groups. It can be observed that the higher weight of the demonstrations are used for imitation learning, the higher the performance.

	Intuitively, high quality demonstration can improve the performance of imitation learning, which implies that the LfND approach can accurately distinguish the quality of different demonstrations by estimating its potential utility. In other words, these results also validate that the LfND approach can effectively emphasize more helpful demonstration while filter out noisy ones.

	\subsection{Study on different weighting strategies }
	As discussed previously, we may weight the instances in different forms. In this subsection, we further compare the results of our method with four different weighting strategies: \textbf{oneZero} denotes the one-zero form of weight as in Eq~\ref{Eq:loss2}; \textbf{linear10} and \textbf{linear20} denote the linear forms of weight with hyperparameter of 10 and 20 respectively as in Eq~\ref{Eq:weight_deta}; \textbf{log} denotes the logarithmic form of weight as in Eq~\ref{Eq:weight_log}.
	
	In Figure~\ref{fig.performanceComparison2}, we plot the reward curves in HalfCheetah, Humanoid and Walker2d environments for six methods: PPO, TRPO, LfND-oneZero, LfND-linear10, LfND-linear20, LfND-log. Firstly it can be observed that LfND with all weighting strategies can achieve better performance than the base RL algorithms. LfND-log, LfND-linear10 and LfND-linear20 outperforms LfND-oneZero in all cases. One intuitive reason is that both linear weight form and log weight form can emphasize the more helpful demonstrations while filter out noisy ones. These results also validate that the superiority of LfND shown in previous experiments is from the proposed strategy of adaptively exploiting noisy demonstrations.
	

	\section{Conclusion}
	\label{section.conclusion}
	In this paper, we propose a new learning framework called LfND to learn policy from noisy demonstrations. With a adaptive weighting strategy to estimate the potential utility of each demonstration instance, the LfND method can learn the policy in a more effective and robust way by incorporating environment exploration and demonstration exploiting. Experiments results with multiple state-of-the-art reinforcement learning algorithms and different environments consistently demonstrate the superiority of the proposed approach. In the future, in addition to the expected gain of value functions, we plan to study other approaches for estimating the potential utility of demonstrations.

	
	\bibliographystyle{named}
	\bibliography{kdd20}

\end{document}